\documentclass[12pt,a4paper]{article}
 
\usepackage[pdftex]{graphicx}
\usepackage[cmex10]{amsmath}
\usepackage{amsfonts}
\usepackage{amssymb}
\usepackage{amsthm}

\usepackage{algorithm}
\usepackage{algpseudocode}
\usepackage{cite}
\usepackage{color}
\usepackage{authblk}
\usepackage{enumerate}
\usepackage{subfigure}
\usepackage{multirow}
\usepackage{rotating}
\usepackage{makecell}
\usepackage{lineno}
\usepackage[font=footnotesize ]{caption}
\usepackage{booktabs}
\usepackage{siunitx}

\usepackage{anysize}
\marginsize{2cm}{2cm}{2cm}{2cm}

\title{Predictive Modeling of Die Filling of the Pharmaceutical Granules Using the Flexible Neural Tree}
\author[1]{Varun~Kumar~Ojha\thanks{Corresponding author~\\Neural Comput \& Applic (2016)~\\DOI:10.1007/s00521-016-2545-8}}
\author[2]{Serena~Schiano}
\author[2]{Chuan-Yu~Wu}
\author[1]{V\'{a}clav~Sn\'{a}\v{s}el}
\author[3]{Ajith~Abraham}
\affil[1]{IT4Innovations, V{\v{S}}B-Technical University of Ostrava, Ostrava, Czech Republic}
\affil[2]{Dept. of Chemical and Process Engineering, University of Surrey, Guildford, GU2 7XH, UK}
\affil[3]{Machine Intelligence Research Labs (MIR Labs), Auburn, WA, USA}
\date{}
\begin{document}
	\maketitle
	\begin{abstract}
		In this work, a computational intelligence (CI) technique named flexible neural tree (FNT) was developed to predict die filling performance of pharmaceutical granules and to identify significant die filling process variables. FNT resembles feedforward neural network, which creates a tree-like structure by using genetic programming. To improve accuracy, FNT parameters were optimized by using differential evolution algorithm. The performance of the FNT-based CI model was evaluated and compared with other CI techniques: multilayer perceptron, Gaussian process regression, and reduced error pruning tree. The accuracy of the CI model was evaluated experimentally using die filling as a case study. The die filling experiments were performed using a model shoe system and three different grades of microcrystalline cellulose (MCC) powders (MCC PH 101, MCC PH 102, and MCC DG). The feed powders were roll-compacted and milled into granules. The granules were then sieved into samples of various size classes. The mass of granules deposited into the die at different shoe speeds was measured. From these experiments, a dataset consisting true density, mean diameter (d50), granule size, and shoe speed as the inputs and the deposited mass as the output was generated. Cross-validation (CV) methods such as 10FCV and 5x2FCV were applied to develop and to validate the predictive models. It was found that the FNT based CI model (in the cases of both CV methods) performed much better than other CI models. Additionally, it was observed that process variables such as the granule size and the shoe speed had a higher impact on the predictability than that of the powder property such as d50. Furthermore, validation of model prediction with experimental data showed that the die filling behavior of coarse granules could be better predicted than that of fine granules.~
	\end{abstract}
	
	\section{Introduction}
	\label{intro}
	In pharmaceutical industries, it is well recognized that the variation in composition and the quality of tablets are determined by material properties and process conditions. One of the greatest challenges in pharmaceutical development is to identify the causal relationship between material properties, intermediate properties, final product properties, and process variables, which is crucial to obtain high-quality products. However, it is a challenging multifactorial problem. 
	
	Tablets are manufactured by compressing dry powders or granules in a die, i.e., the so-called die compaction. The die compaction process consists of three primary stages: die filling, compaction, and ejection~\cite{coube2005experimental}. Simple gravity effect can play a role during die filling of powders into the die. However, flow behavior during die filling process controls the tablet composition, the tablet properties as well as its segregation tendency~\cite{wu2003flow,schneider2007characterisation,wu2008simulations}. Therefore, the study of die filling process parameters has a significant role in controlling tablet (drug) manufacturing industry.
	
	Previous studies on die filling have been focused on such factors influencing the flow behavior as powder characteristics, apparatus features, and operating variables. For instance, the effect of density and particle size on gravity and suction filling of several grades of microcrystalline cellulose powders were explored in~\cite{mills2013effect}. It was shown that fine particles present intermittent flow behavior due to their cohesive property, while larger particles are free flowing. The suction effect during die filling on a rotary press system was studied in~\cite{jackson2007effect}, where the strong influence on the flow behavior of powders was observed due to air pressure build-up into an open cavity. Moreover, it was found that the lower punch has to be withdrawn at a higher speed by increasing the turret speed in order to improve the suction feed efficiency.  
	
	In~\cite{lawrence1968some}, authors studied powder segregation behavior during die filling process for a mixture of fine and coarse powders from a stationary hopper. Segregation was found due to the movement of the fine particles on the powder bed during the process. Moreover, the reduction of segregation was possible with the decrease of fill time or enhancing the drop height. 
	
    Several studies reported in the literature are on apparatus features such as die width and feed frame design. For example, a study on die width proved that the filling density decreases proportionally with the die width~\cite{bocchini1987influence}. In another study, the filling density was found correlated with the shoe speed, i.e., the filling density was observed to be increasing as the shoe speed was decreasing~\cite{rice1986filling}. Additionally, in ~\cite{mendez2012powder}, authors studied the influence of different feed frames of powder blends and found that the feed frame had a significantly high impact on powder hydrophobicity due to over lubrication of the powders caused by the shear. This will lead to an increase of flowability and a decrease in tablet tensile strength.
	
	Similarly, different process conditions such as die filling in the air or in the vacuum, were studied~\cite{wu2003flow}, where it was shown that nose flow and bulk flow are two possible classifications of die filling depending on the shoe speed, the powder flow patterns. Moreover, a faster filling rate was observed in the vacuum than the filling rate in the air. Furthermore, in~\cite{wu2004flow}, authors proposed a new methodology based on the concept of critical shoe velocity to characterize powder flow, where a critical fill speed that suggests the minimum filling speed at which the die can be filled, completely. The die will not be full if the velocity is higher than the critical velocity. Hence, it was possible to determine the fill ratio at these higher speeds using the following equation:
	\begin{equation}
	\label{eq_}
	\delta  = \left(\frac{v_c}{v_s} \right)^m	
	\end{equation}				
	where $ v_c $ is the critical velocity, $ v_s $ is the shoe speed, and m is a coefficient of value between 1.0 and 1.6. 
	
	Guo et al.~\cite{guo2009coupled} used a coupled discrete element method with computational fluid dynamics (DEM-CFD) to modeled die filling process, in which they explored the influence of powder properties (i.e., particle size and density) on die filling in the air and in the vacuum. They characterized flow in terms of mass flow rate and found that the mass flow rate is constant in the vacuum; whereas for light and small particles in the air-sensitive regime, it was found that the mass flow rate increases as the particle size and density increase, and for coarse and dense particles in the air-inert regime, a negligible effect on airflow was observed. Guo et al.~\cite{guo2011effects} then investigated segregation induced by air during die filling, where DEM-CFD was used to simulate monodispersed mixtures. For die filling using a stationary shoe, segregation was induced due to the air presence and a lower concentration of light particles was found in the lower region of the die. Whereas, for die filling using a moving shoe, lighter particles predominantly gather in the leading edge of the flow stream. 
	
	Although several investigations on die filling have been proposed both experimentally and numerically, the influence of the granule property on die filling behavior and how this is correlated with the raw material properties and process variables are not fully understood. Moreover, it is of commercial benefit for pharmaceutical companies to reduce the development time and costs with the contemporary improvement of the pharmaceutical process design~\cite{zhao2006toward}. To achieve these goals, the use of the multilayer perceptron (MLP) on a limited set of experiment data for predictive modeling can be of great benefit~\cite{bourquin1998advantages}. Additionally, use of computational intelligence techniques is advantageous for various pharmaceutical processes~\cite{wu2002optimal,kim2000application}. Hence, the objective of this work is to develop a predictive model that can accurately describe the die filling behavior of pharmaceutical granules. For this purpose, a computational intelligent (CI) technique, named flexible neural tree (FNT), is used for predicting die filling performance of pharmaceutical granules
	
	\section{Computational intelligence techniques}
	\label{sec_die_ci_methods}
	Computational intelligence techniques discover knowledge from data and create predictive models. Moreover, in a predictive modeling, a causal relationship between the independent variable $ X = (\text{\bf{x}}_1,\text{\bf{x}}_2,\ldots,\text{\bf{x}}_N) $ and the dependent variable $ \text{\bf{d}}= \langle d_1,d_2,\ldots,d_N \rangle $ is discovered. Specifically, the unknown parameter $ \text{\bf{w}} $, which represents the learning component of a CI technique, are determined by minimizing the root mean square error $ e $ (RMSE) between the desired output $ \textbf{d} $ and the predicted output $ y = \langle y_1,y_2,\ldots ,y_N \rangle $. Hence, the training of a CI technique is a process of searching the optimum  parameter $ \text{\bf{w}} $. In this work, the RMSE $ e $ was computed as:
	\begin{equation}
	\label{eq_rmse}
		e= \sqrt{ \frac{1}{N} \sum_{i=1}^N \left( y_i - d_i \right)^2},
	\end{equation}
	where $ N $ is total samples in a training set. An additional performance measure used in this work was correlation coefficient $ r $ that quantifies the correlation between two the variables: the desired output $ d $ and the predicted output $ y $. The correlation coefficient value $ r $ was computed as:
	\begin{equation}
	\label{eq_corr}
        r = \frac{\sum_{i=1}^{N}((y_i - \text{mean} (y))(d_i - \text{mean}(d)))}{\sum_{i=1}^{N}(y_i - \text{mean}(y))^2\sum_{i=1}^{N}(d_i - \text{mean}(d))^2}    
	\end{equation}	
	where function $ \text{mean}(\cdot) $ returns the average of the elements of a vector. The correlation coefficient $ r $ equal to 1 indicate the best performance and the correlation coefficient $ r $ equal to -1 indicates the worst performance. 
	
	The common CI techniques such as multilayer perceptron (MLP), Gaussian process regression (GPR), and reduced error pruning tree (REP-Tree) solves several industrial and real-world problems~\cite{lam2012computational}. The concise definitions of these CI techniques are as follows:
	\begin{itemize}
		\item The MLP is a mathematical form of human-like learning, where a network of computational nodes arranged in a layered architecture is trained using a dataset~\cite{haykin2009neural}.
		\item In REP-Tree, based on the dataset, a tree-like structure is created, where the tree's internal nodes are binary decision nodes and the leaf nodes are the output nodes~\cite{kohavi2002data}.
		\item The GPR is a Gaussian distribution based extension of linear regression technique~\cite{rasmussen2006gaussian}. 
		 
	\end{itemize}
	These CI techniques are implemented as an open source library, i.e., as a software tool named WEKA~\cite{weka2016}. A detailed description of these models is available in~\cite{hall2009weka}. 
		
	In comparison to the models above, the flexible neural tree (FNT) produces a tree-like model, where the nodes of the tree are similar to the MLP nodes. Hence, FNT differs from MLP in its structural configuration, and it differs from REP-Tree in its node type. Moreover, the tree-like structure in the FNT is created using genetic programming (GP)~\cite{chen2004nonlinear,chen2005time}. Therefore, the basic advantage of using FNT over MLP and REP-Tree lies in its ability of automatic adaptation into the tree-structure and the input feature selection with the help of GP~\cite{poli2008field}. Such an ability is necessary and has been used in several successful applications. Examples of such applications are cement decomposing-furnace production-process modeling~\cite{shou2008modeling}; exchange rate forecasting~\cite{chen2008foreign}; gene regulatory network reconstruction and time-series prediction from gene expression~\cite{yang2013reverse}; Internet traffic identification~\cite{chen2015flexible}; and protein desolation rate prediction~\cite{ojha2016ensemble}.
	
	It is known that for a good quality data, the predictive capability of MLP lies in its learning components, such as connection weights, network-architecture, activation-function, input features, and learning rules~\cite{yao1999evolving}. The design of FNT includes all these MLP-like learning components and offer an automatic adaptation of these components by using GP~\cite{chen2004nonlinear,chen2005time}. The descriptions of the CI models are given below.
	
	\subsection{Multilayer Perceptron (MLP)}
	Usually, an MLP structure has three layers: an input layer, a hidden layer, and an output layer. Each layer of an MLP consists of several nodes, and each node receives real-valued inputs and produces outputs based on the weighted linear combination of inputs. Then, the magnitude of the node's output is computed by using a nonlinear activation function. In this work, resilient propagator algorithm was used to train an MLP~\cite{riedmiller1993direct}.
	
	\subsection{Gaussian process for regression (GPR)}
	GPR is a statistical technique that extends multivariate Gaussian distribution. Let $ \hat{y} = f(x) = \phi(\text{\bf{x}})^T \text{\bf{w}} $  be a regression model, where an input vector $ \text{\bf{x}}= \langle x_1,x_2,\ldots ,x_d \rangle  \in X  $ is mapped onto an $ N $-dimensional feature space $ \phi(\cdot) $, and $ \text{\bf{w}} $ is the learning component. Then, a mean function $ m(x) $ and a covariance function $ k(x,x^{'}) $ defines a Gaussian process $ gp(\cdot) $. The mean and covariance functions are written as:
	\begin{flushleft}
	$ m(x)= \mathbb{E} [f(x)], $~\\
	$ k(x,x^{'} )=\mathbb{E}[(f(x)-m(x)),(f(x^{'} )-m(x^{'} ))]. $~\\~\\
	and Gaussian process $ gp(\cdot) $ may be written as: ~\\
	$ f(x) \sim gp(m(x),k(x,x^{'})). $
	\end{flushleft}
    The mean function $ m(x) $ and the covariance function $ k(x,x^{'}) $ are described in~\cite{rasmussen2006gaussian}.
	
	\subsection{Reduced error pruning tree (REP-Tree)}
	Reduced error pruning tree depends on the creation of a decision tree. A decision tree resembles a tree-like structure whose non-leaf nodes are decision nodes. A decision node is tested (a binary decision) against an input feature and the leaf node are the outcomes of the test. In REP-Tree, a decision tree is created using the information available in a training data and then the leaves and branches of the tree are pruned using the reduced-error pruning method to avoid over-fitting. In the reduced error pruning method, the nodes of the tree are pruned until the accuracy of the prediction is not affected. A detailed study of this algorithm is available in\cite{kohavi2002data}.
	
	\subsection{Flexible neural tree (FNT)}
	To minimize RMSE $ e $ so that it converges to zero, an FNT adapts its parameter by using learning algorithms (such as genetic programming and differential evolution). Such adaptation is performed by optimizing the components of tree-like structure: internal-nodes (computational nodes), branches (connection-weights), and leaf nodes (terminal nodes). 
	
\begin{figure}
		\centering
		\includegraphics[width=0.6\textwidth]{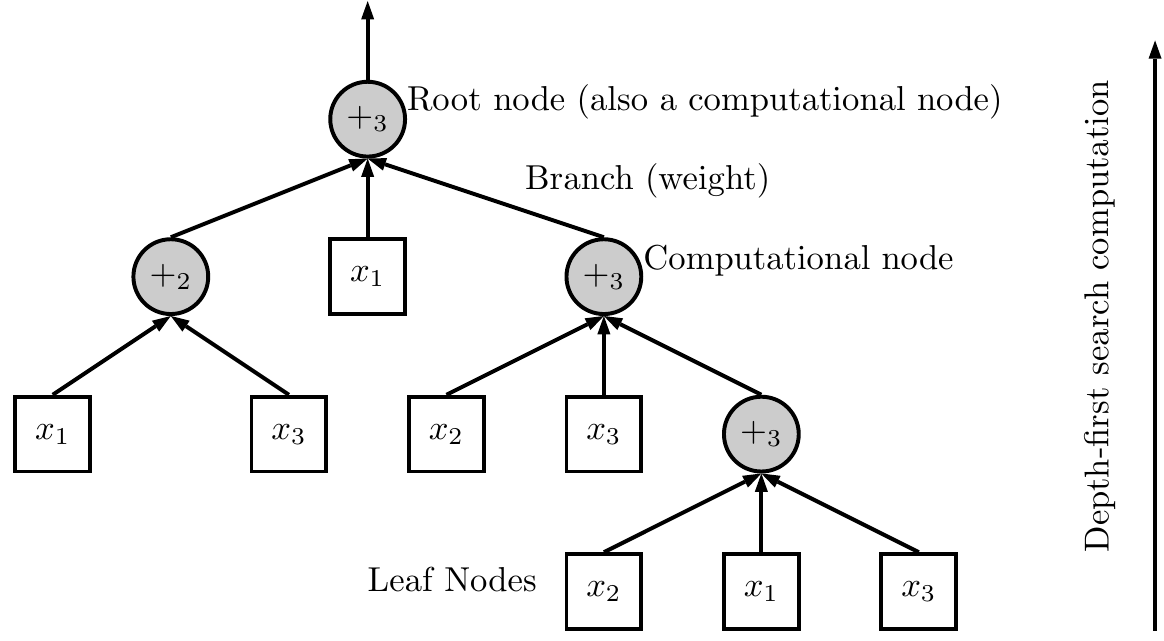}
		\caption{A typical FNT $ S $ with computational set $ F= \{ +_2,+_3\} $ and terminal set $ T={x_1,x_2,x_3} $.}
		\label{fig_fnt_gen}
	\end{figure}

	Mathematically, a union of computational node $ F $ and the terminal node $ T $ describes an FNT denoted as $ S $, where a computational node represents an activation function, and a terminal node represents an input. Hence, an FNT can be represented as:
	\begin{equation}
	\label{eq_fnt}
	S = F \cup T =  \{+_2,+_3,\cdots,+_N \} \cup \{x_1,x_2, \ldots, x_d \}
	\end{equation}
	where $ +_n $ ($ n=2 $ or $ 3 $ or $ \ldots $ or $ N $) indicates that a computational node that takes two or more arguments; whereas, the terminal node takes zero arguments. Fig.~\ref{fig_fnt_gen} is an illustration of FNT model. 
		
\begin{figure}
	\centering
	\includegraphics[width=0.4\textwidth]{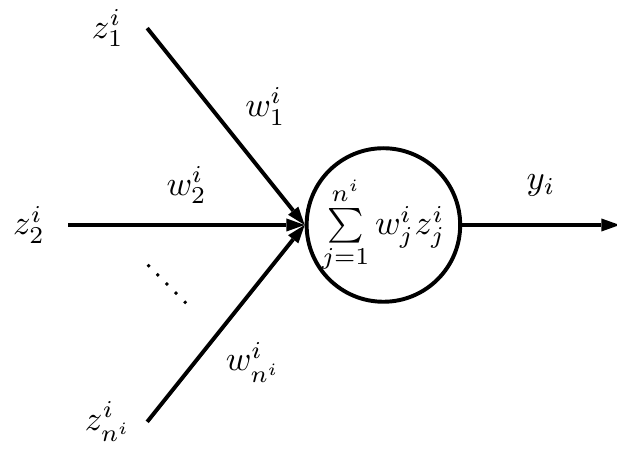}	
	\caption{Illustration of an FNT's computational node. The variables $ z_j^i $ and $ w_j^i $ in the figure indicate the input and the weight. The variable $ y_i $ is the output of the node.}
	\label{fig_fnt_node}
    \end{figure}

	The shaded circle in Fig.~\ref{fig_fnt_gen} represents a computational node (the root node is among the computational nodes), the branches of the tree indicate the weights of the model, and the leaf nodes (shown in square) of the tree indicates the selected input features. The root node of the tree represents the output of the model.The output of an FNT at the root node is computed using a bottom-up approach (as indicated by arrows in Fig.~\ref{fig_fnt_gen}).
	The $ i $-th computational node (shown in Fig.~\ref{fig_fnt_node}) of the tree (Fig.~\ref{fig_fnt_gen}) receives $ n^i $  input through $ n^i $ connection-weights (branches) and two adjustable parameters $ a_i $ and $ b_i $, which represent the arguments of the $ i $-th computational node's activation function. The purpose of using an activation function at the computational node is to limit computational node's output within a certain range. For example, output of $ i $-th node that contains a Gaussian function is computed as:
	\begin{equation}
	\label{eq_fnt_y}
	y_i= f_i (a_i,b_i,o_i )= \exp \left(-\left(\frac{(o_i-a_i)}{b_i} \right)\right)			
	\end{equation}
	where $ o_i $ is the weighted summation of the inputs to the $ i $-th computational node (see Fig.~\ref{fig_fnt_gen}), which is computed as:
	\begin{equation}
	\label{eq_node_o}
	o_i= \sum_{j=1}^{n^i} w_j^i x_j^i	
	\end{equation}
	where $ z_j^i \in \{x_1,x_2,\ldots,x_d\} $ or  $ z_j^i \in \{y_1,y_2,\ldots,y_k\} $, i.e., $ z_j^i $ can be either an input feature (leaf node value) or the output of another node (computational node value). The weight $ w_j^i $ is the connection weight of real value in the range $ [w_l,w_h] $. Now, two aspects are involved in the training of an FNT: tree structure optimization and parameter optimization. The optimization of these two aspects of FNT took place in two phases: 
	\begin{enumerate}[1)]
		\item Structure optimization using genetic programming (GP) 
		\item Parameter optimization with differential evolution (DE). 
	\end{enumerate}
	
	\subsubsection{Structure optimization phase}
	GP is a population-based evolutionary-inspired algorithm that makes use of genetic operators such as crossover and mutation and iteratively constructs an optimal tree-like structure (the structure that gives the lowest approximation error) through a vast topological search space~\cite{poli2008field}. In a crossover operation, the randomly selected subtrees of two trees are exchanged between each other's and in a mutation operation a randomly selected subtree is either deleted entirely or replaced by a randomly generated subtree. In this work, the  mutation operators were designed as follows:
	\begin{enumerate}[1)]
		\item Replacing one terminal node with a new terminal node;
		\item Replacing all terminal nodes with new terminal nodes;
		\item Growing a new sub-tree and replacing a randomly selected function node;
		\item Replacing a computational node with a terminal node.
	\end{enumerate}
	Since GP is an evolutionary-inspired algorithm, only the best input features (regarding the predictability) are selected during the structure optimization phase.
	\subsubsection{Parameter optimization phase}
\begin{figure}
		\centering
		\includegraphics[width=0.8\textwidth]{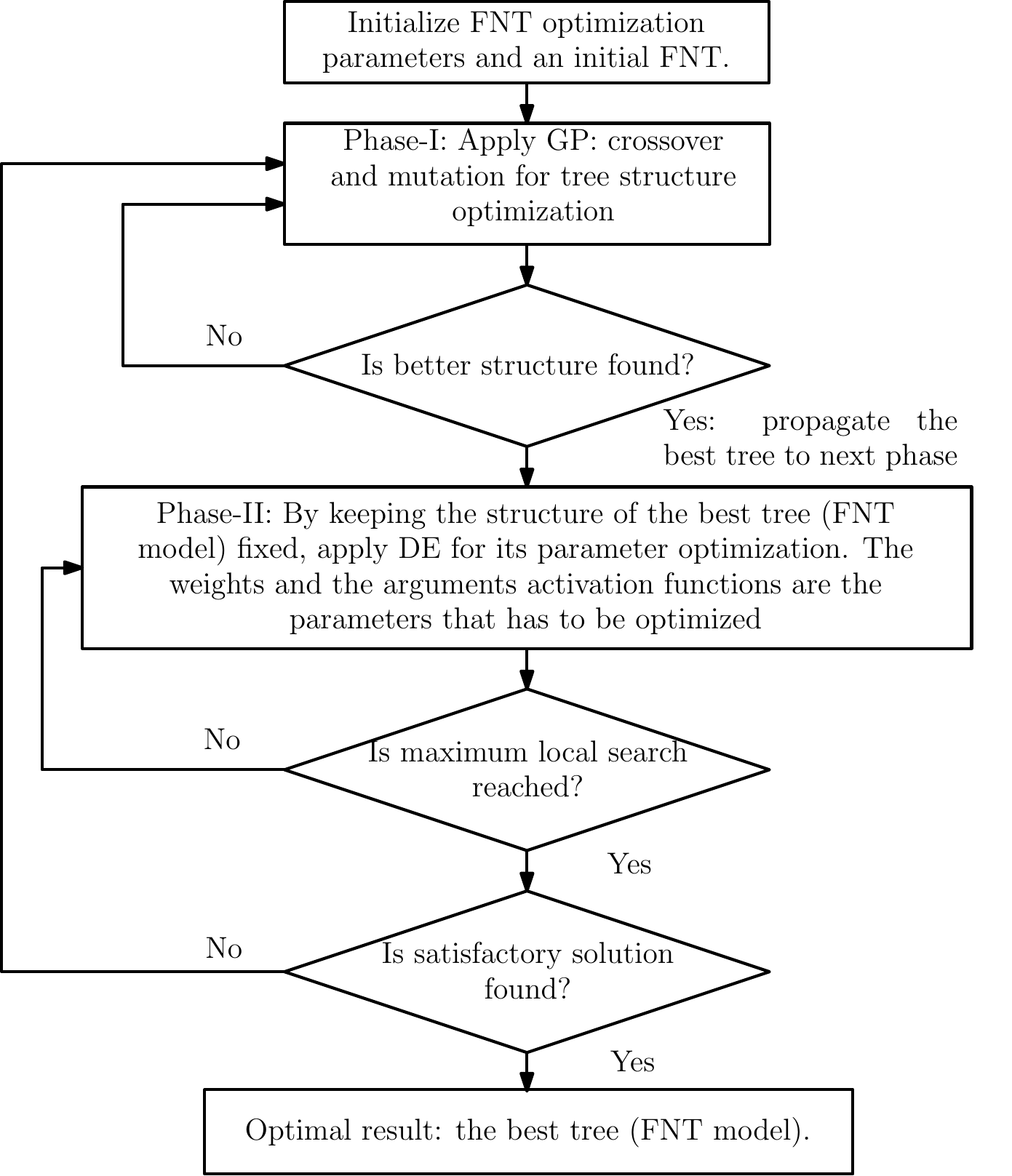}
		\caption{General FNT optimization procedure}
		\label{fig_fnt_flow}
	\end{figure}

	DE is a population-based parameter optimization algorithm that makes use of crossover operator and iteratively searches through a search space to optimize a vector that represents the parameters of a model~\cite{storn1997differential}. Let assume that $ H $ is the population of the tree parameter vectors. Hence, DE tries to find an optimal tree parameter $ \text{\bf{h}}^{*} $  by using operators, such as selection, crossover, mutation, and recombination. At each iteration, the selection operator selects three random vectors from the population $ H $: $ \text{\bf{r}}_1 $, $ \text{\bf{r}}_2 $, and $ \text{\bf{r}}_3 $ such that $ \text{\bf{r}}_1 \ne \text{\bf{r}}_2 \ne \text{\bf{r}}_3 $ and the vector $ \text{\bf{g}} $ that denotes the best solution in the population. Therefore, a new tree parameter vector $ \text{\bf{h}}^{new} $ is computed as:
	\begin{eqnarray}
         \text{\bf{h}}^{new} \begin{cases} \text{\bf{r}}_1 + F \cdot (\text{\bf{r}}_1 - \text{\bf{g}}) + F \cdot (\text{\bf{r}}_2 - \text{\bf{r}}_3), & \mbox{if } \text{\bf{u}} < \text{\bf{C}} \\ \text{\bf{r}}_1, & \mbox{if } \text{\bf{u}} \ge \text{\bf{C}}  \end{cases}
	\end{eqnarray}
	where $ \text{\bf{u}} $, is a vector of random values taken from a uniform distribution between 0 and 1, \textbf{C} is a vector of crossover probability, and $ F $ is a vector of mutation factor. This process is repeated until $ \text{\bf{h}}^* $  is found. Hence, by combining structure optimization phase and parameter optimization phase, a general optimization procedure of FNT can be summarized in Fig.~\ref{fig_fnt_flow}.
	
	\subsubsection{Limitations of FNT}
	The fundamental limitation of the FNN is in comparison to MLP is at its output node, i.e., the root node of the FNT (tree-like structure). Since the tree can have only one root node, one has to produce multiple models for multiple output prediction problems. Moreover, FNT can only offer multi-input-single-output models; whereas MLP can provide multi-input-multi-output models. Additionally, the proposed improvement in FNT in comparison to other CI techniques is its structure optimization. However, during cross-validation (CV) methods such as $ k $-fold CV, the structure is determined only once (for only first fold) and for the rest of the ($ k-1 $) folds only parameters were optimized. However, if the structure is determined for all the $ k $ folds. Then, $ k $ different FNT models should be obtained. 
	
	\section{Experimental study}
    \subsection{Materials and methods}
	Microcrystalline cellulose (MCC) of three different grades was chosen as the raw materials, including Avicel PH-101, Avicel PH 102, and DG. A custom-made gravity-fed roll compactor with two counter rotating smooth rolls of 200~\si{\milli\meter} in diameter and 46~\si{\milli\meter} in width was used for ribbon production~\cite{zhang2016application,schiano2016novel}. The roll gap was set at 1.2~\si{\milli\meter} and the roll speed at 1 rpm roll speed. Ribbons were milled using a milling system (SM100, Retsch, Germany) equipped with a mesh size of 4~\si{\milli\meter} at a constant speed of 1,500 rpm. The granules were sieved into different size classes (0-90, 90-250, 250-500, 500-1000, 1000-1400, 1400-2360~\si{\micro\meter}) that were used for the die filling experiments. The corresponding upper size limit was used as the granule size in the current study (i.e., granules in the size range 0-90~\si{\micro\meter} were regarded as granules with a size of 90~\si{\micro\meter}), as commonly adopted in particle technology.
	\subsection{Data collection}
	True densities of the three MCC powders were determined using a Helium Pycnometer (AccuPyc II 1340, Micromeritics, UK). Particle size analysis was performed using a size analyzer (Camsizer XT, Retsch, UK). The experiments were run for 2-3 minutes and the data were collected. The mean diameter (d50) defined as the size value below which 50\% of the particles lies, was then determined. Die filling experiments were performed using a model die filling system that consists of a shoe driven by a pneumatic driving unit, a positioning controller, and a displacement transducer~\cite{wu2003flow}. For each granule size class, experiments were performed using seven different shoe speeds in the range of 10~\si{\milli\meter\per\second} to 400~\si{\milli\meter\per\second}. For each die filling experiment, the powder mass deposited into the die was weighted, and the value was recorded. Each experiment was repeated three times. In total, 389 experiments (3 powders, 6 granule sizes, 7 speeds, 3 repeats) were performed, and 389 datasets were generated.
	
	From the collected experimental data, four parameters were chosen as inputs for the modeling: true density and mean diameter (d50) of raw powders, granule size (~\si{\micro\meter}), and shoe speed (~\si{\milli\meter\per\second}) and the deposited mass was the only output. Table~\ref{tab_die_data} shows a selection of few samples (taken from 389 samples) of the generated dataset.
	
	\begin{table}
		\centering
		\caption{Example of few data samples generated for modeling.}
		\label{tab_die_data}
		\footnotesize
		\begin{tabular}{lllllll}
			\toprule
			\multicolumn{2}{c}{Samples}  & \multicolumn{4}{c}{Input} & Output\\
			\cline{3-6}
			\# & Name & True density & d50 (~\si{\micro\meter}) & Granules size & Shoe speed & Mass (g)\\
			\cline{3-6}
			&  & Feature \#1 & Feature \#2 & Feature \#3 & Feature \#4 & \\
			\midrule
			1 & MCC PH 101 & 1581 & 59.83 & 90 & 10 & 12.81\\
			2 & MCC PH 101 & 1581 & 59.83 & 90 & 10 & 12.78\\
			: & : & : & : & : & : & :\\
			5 & MCC PH 101 & 1581 & 59.83 & 90 & 20 & 12.3\\
			6 & MCC PH 101 & 1581 & 59.83 & 90 & 30 & 9.55\\
			: & : & : & : & : & : & :\\
			135 & MCC PH 102 & 1570.3 & 94.7 & 250 & 50 & 13.45\\
			136 & MCC PH 102 & 1570.3 & 94.7 & 250 & 60 & 13.5\\
			: & : & : & : & : & : & :\\
			388 & MCC DG & 1785.6 & 52.33 & 2360 & 400 & 9.51\\
			389 & MCC DG & 1785.6 & 52.33 & 2360 & 400 & 9.3\\
			\bottomrule
		\end{tabular}		
	\end{table}	 
	\subsection{Predictive modeling}
\begin{figure}
		\centering
		\includegraphics[width=0.8\textwidth]{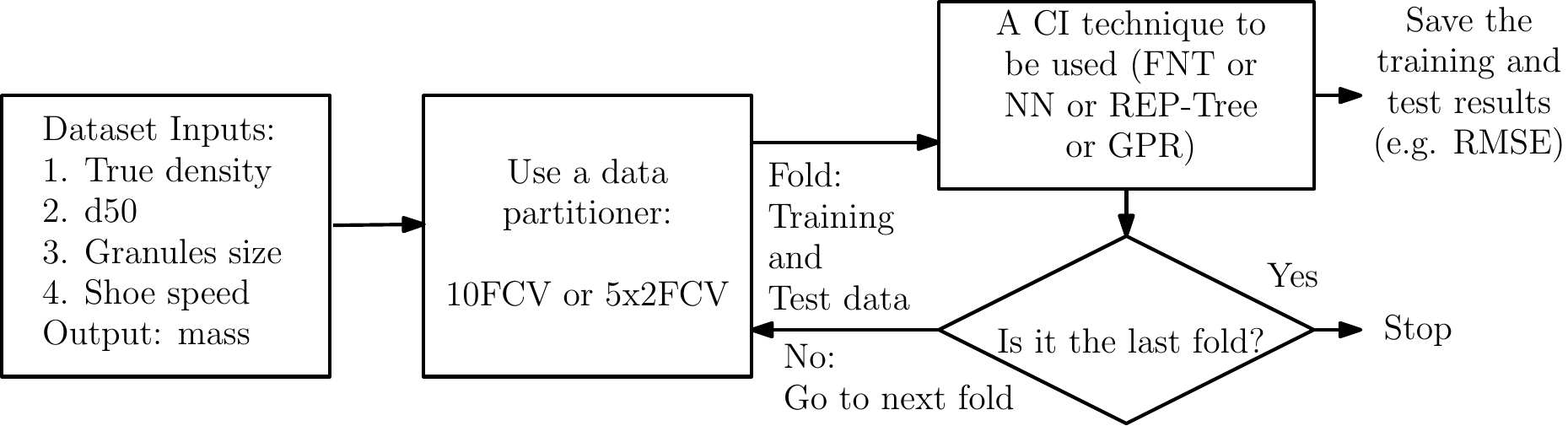}
		\caption{Flow chart illustrating the data flow during a predictive modeling.}
		\label{fig_die_data_flow}		
	\end{figure}

    Predictive modeling with the CI techniques discussed in Section~\ref{sec_die_ci_methods} was performed in the following manner. Two different cases, i.e., 10-fold cross validation (10FCV) and 5-cross-2-fold cross validation (5x2FCV) methods, were considered. In 10FCV, a dataset is first randomized (shuffled). Then, 90\% of data samples are used for training of a model and the rest of 10\% data samples are used for testing. This process is repeated for 10 times and each time a distinct set of 10\% dataset is picked for testing of the model. Hence, called 10FCV. In the 5x2FCV, the data set is randomly divided into two sets, where each set is 50\% in size of the entire dataset. When the first set is used for the training of a model, then the second set is used for the testing of the same model, and the vice-versa. This process is repeated for five times hence referred to as 5x2FCV validation. In each case, the four CI techniques discussed above (i.e., FNT, MLP, GPR, and REP-Tree) were used to develop the models, and their performance was evaluated. Fig.~\ref{fig_die_data_flow} illustrates the predictive modeling process adopted in this work. 
	
	The collected dataset was pre-processed by normalizing the features of dataset between zero and one using a min-max normalization method. Table~\ref{tab_fnt_para} lists the underlying parameters for the FNT, whereas, Table~\ref{tab_ci_model_para} describes the basic parameter set-up for other three CI techniques: MLP, GPR, and REP-Tree. The model accuracy was assessed using RMSE $ e $ defined in~\eqref{eq_rmse} and correlation coefficient $ r $ defined in~\eqref{eq_corr}.	
	
	\begin{table}
		\centering
		\caption{Parameters settings and the values are chosen during FNT optimization.}
		\label{tab_fnt_para}
		\footnotesize
		\begin{tabular}{llll}
			\toprule
			\# & Parameter Name & Definition/Purpose & Value\\
			\midrule
			1 & Tree height & The maximum levels of a tree. & 5\\
			2 & Tree arity & Maximum siblings of a node in a tree. & 4\\
			3 & Tree node type & Type of activation function used at nodes. & Gaussian\\
			4 & GP population & Total candidates in a GP population. & 30\\
			5 & Mutation probability & The frequency of mutation operation. & 0.2\\
			6 & Crossover probability & The frequency of crossover operation. & 0.8\\
			8 & Tournament size & Total candidates for a tournament. & 15\\
			9 & DE population & The initial size of the DE population. & 50\\
			10 & Range of node search  & The lower and upper bound of activation function arguments. & [0.0,1.0]\\
			11 & Range of edge search & The lower and upper bound of tree edges. & [-1.0,1.0]\\
			13 & Structure training & Maximum generations of GP training. & 100~000\\
			14 & Parameter training & Maximum evaluations of parameter training. & 10~000\\
			\bottomrule
		\end{tabular}		
	\end{table}	
	
	\begin{table}
		\centering
		\caption{Parameters settings and the values are chosen during MLP, GPR, and REP-Tree training. The settings mentioned are those used in the software tool~\cite{weka2016}}
		\label{tab_ci_model_para}
		\footnotesize
		\begin{tabular}{lllp{7cm}l}
			\toprule
			\# & CI Technique & Parameter Name & Definition/Purpose & Values \\
			\midrule
			1 & MLP & Learning rate & Convergence speed. & 0.3\\
			2 &  & Momentum rate & Magnitude of past iteration influence. & 0.2\\
			3 &  & Hidden Layer & Maximum nodes at hidden layer. & 350\\
			4 &  & Iterations & Maximum number of evaluations of parameter optimization & 500\\
			\midrule
			5 & GPR & Kernel & The function used for implementing covariance function. & RBF Kernel\\
			\midrule
			6 & REP-Tree & No. Leaf Instances & Minimum No. of children per node. & 2\\
			8 &  & Depth & Maximum limit of tree depth/level. & No limit\\
			9 &  & Pruning & Pruning of tree nodes. & Allowed\\
			\bottomrule
		\end{tabular}		
	\end{table}
	To obtain the best model, the training and test performances of the models were observed using five performance indicators: 
	\begin{enumerate}[1)]
		\item The RMSE $ e $ as per~\eqref{eq_rmse}. 
		\item The correlation coefficient $ r $ as per~\eqref{eq_corr}.
		\item The standard deviation (std) of correlation coefficient values of each fold.
		\item The model's complexity. In the case of tree-based models (FNT and REP-Tree), it is the total number of nodes in the tree including computational node and leaf nodes (e.g., the complexity of the model shown in Fig.~\ref{fig_fnt_gen} is 12). In the case of MLP, it is the number of total computational nodes. Since GPR is an extension of the linear regression model, the complexity in the case of GPR was not commuted.
		\item The number of selected input features.  	
	\end{enumerate}		
   
    \subsection{Input feature analysis of die filling experiments}
    Feature analysis was conducted to understand the significance of the input features in the die filling experiments. For this purpose, at first, $ M $ many FNT models were created. Second, two performance dimensions were used: feature selection rate $ R $ and predictability score $ P $. Feature selection rate $ R $, which describes the total number of times a particular input feature set was appeared in the list of prepared out all the M models. The feature selection rate is defined as:
    \begin{equation}
    \label{eq_sel_rate}
    R_j =  \frac{1}{M} \sum_{i=1}^M \mathbb{I}(A_j) 
    \end{equation}
	where $ R_j $ is the selection rate of $ j $-th input feature set $ A_j \in \text{\bf{A}} $ which is a power set $ P\{$true density, d50, granule size, shoe speed$\}$,  $ M $ is the total number of models in the list, and function $ \mathbb{I}(A_j) $ is an identity function that returned ``1" if $ j $-th input feature set $ A_j $ is selected by the $ i $-th model, otherwise, returned ``0." Feature selection rate $ R $ equal to one is the highest (100\% selection rate), and $ R $ equal to zero is the lowest (0\% selection rate). In other words, the value of selection rate $ R $ equal to one means an input feature was selection by all the models in the prepared list and the value of $ R $ equal to zero means an input feature was selection by none of the models in the prepared list.
	
	Since in the models in the list may not equal in their performances, the predictability score P based on the RMSEs of the models was computed. The predictability score $ P $ describes the predictability of an input feature. To compute predictability score $ P $ of $ j $-th input feature set $ A_j $, at first, the fitness $ F_j $  of the corresponding input feature set $ A_j $  was computed as:
	\begin{eqnarray}
	\label{eq_pred_score}
	F_j = \begin{cases} \sum_{i=1}^{M} e_i \cdot \mathbb{I}(A_j), & \mbox{if } |A_j| = 1 \\ 
	\sum_{i=1}^{M} e_i \cdot \mathbb{I}(A_j) / \sum_{i=1}^{M} \mathbb{I}(A_j), & \mbox{if } \mbox{if } |A_j| \ge 1  \end{cases}
	\end{eqnarray}
	where $ e_i $ is the RMSE of $ i $-th model. The fitness $ F_j $ for $ |A_j| $ equal to one is the sum of RMSEs, and the $ F_j $ for $ |A_j| $ greater than one is the average RMSEs of all models that selects subset $ A_j $. Then, the predictability score $ P_j $ corresponding to an input feature set $ A_j $  was computed by normalizing the fitness of as:
	\begin{equation}
	\label{eq_pred_soce_final}
	P_j = \frac{F_j}{\smash{\displaystyle\max_{j = 1 \text{ to } z}} (F_j)} 
	\end{equation}
	where function $ \max(\cdot) $ determines the maximum fitness value from all $ F_j $. Similar to the selection rate $ R $, the predictability score $ P $ equal to one indicates that the feature set has the highest impact on the predictability of the model and predictability score $ P $ equal to zero has the lowest.
	
	\section{Results and discussion}
	\subsection{Model Prediction}
	\subsubsection{Predictions using the 10FCV method}
	Table~\ref{tab_die_10FCV_res} describes the performance of the best models created by using FNT, MLP, GPR, and REP-Tree. In Table~\ref{tab_die_10FCV_res}, the generated models are arranged in descending order (highest accuracy to lowest accuracy) of their test correlation values. It may be observed that the performance of the models created using FNT (when compared the test accuracies, i.e., correlation values) was better than the other CI techniques. Hence, the detail observation of the model Nos. 1, 2, and 3, which were created using FNT are provided.

	The model No. 1 (see: row 1 of Table~\ref{tab_die_10FCV_res}) produces a high correlation coefficient (on test set), i.e., 0.95, which indicates high predictability of this model over 10\% of unknown samples. However, the model complexity was also high since the total function nodes, and leaf nodes in the created model amounted to 43. In addition, there was no feature selection performed by this model. In comparison to model No. 1, the model Nos 2 and 3 had lower test correlation coefficient (performance was slightly poor). However, their model’s complexity was simpler, and they offered feature selection, which was advantageous than the model No. 1. Fig.~\ref{fig_die_10FCV_res} illustrates the performance of the model using regression (scatter) plot and target against predicted value plot.
	In Fig.~\ref{fig_die_10FCV_res}, each model Nos 1, 2, and 3 respectively were tested over 10\% of test samples, i.e., 38 randomly chosen test samples, and the scattered plot and the target against predicted values plot were analyzed. It may be observed that the prediction curve follows the target curve. However, the lower values and outliers were slightly out of the reach in the prediction curve. 
	
	\begin{table}
		\centering
		\caption{Performance of the prediction models and validation over 10FCV.}
		\label{tab_die_10FCV_res}
		\footnotesize
		\begin{tabular}{llrrrrrrrl}
			\toprule
			Model  & Model  & \multicolumn{2}{c}{Mean of RMSEs}  & \multicolumn{2}{c}{Mean of $ r $} & \multicolumn{2}{c}{Std over $ r $} & Model  & Selected  \\
			\cline{3-8}
			No. &  Type & Train & Test & Train & Test & Train & Test & Complexity$ ^1 $ & Features$ ^2 $ \\
			\midrule
			1 & FNT 	 & 2.0206 & 2.0571 & 0.93 & 0.95 & 0.0087 & 0.0383 & 43 & 1, 2, 3, 4\\
			2 &  		 & 2.3891 & 2.3934 & 0.91 & 0.91 & 0.0083 & 0.0617 & 34 & 2, 3, 4\\
			3 &  		 & 2.5491 & 2.2618 & 0.88 & 0.91 & 0.0078 & 0.0563 & 32 & 3, 4\\
			4 & REP-Tree & 2.5751 & 3.1637 & 0.88 & 0.82 & - & - & 99 & 1, 2, 3, 4\\
			5 & GPR      & 2.9632 & 3.4023 & 0.86 & 0.79 & - & - & - & 1, 2, 3, 4\\
			6 & MLP       & 3.3687 & 3.4427 & 0.81 & 0.79 & - & - & - & 1, 2, 3, 4\\
			\bottomrule
			\multicolumn{10}{l}{\scriptsize\textbf{Note}: $ ^1 $Complexity is the sum of total nodes in the created tree-model. $ ^2 $Features Nos are assigned in Table~\ref{tab_die_data}} \\
		\end{tabular}	
	\end{table}		
	\subsubsection{Predictions using the 5x2FCV method}
	In Table~\ref{tab_die_5x2FCV_res}, a comparison of the best models created using FNT, MLP, GPR, and REP-Tree are provided. The models in Table~\ref{tab_die_5x2FCV_res} are arranged in descending order (the highest accuracy to the lowest accuracy) of their test correlation values. Similar to the modeling in 10FCV, the modeling in 5x2FCV and the performance of the models created using FNT outperformed the models created using MLP, GP, and REP-Tree. Hence, a detail observation on the model Nos. 7, and 8 are provided. Fig.~\ref{fig_die_5x2FCV_res} illustrates the performance of the model using regression (scatter) plot and target against predicted value plot.
	
	\begin{table}
		\centering
		\caption{Performance of the prediction models and validation over 5x2FCV}
		\label{tab_die_5x2FCV_res}
		\footnotesize
		\begin{tabular}{llrrrrrrrl}
			\toprule
			Model  & Model  & \multicolumn{2}{c}{Mean of RMSEs}  & \multicolumn{2}{c}{Mean of $ r $} & \multicolumn{2}{c}{Std over $ r $} & Model  & Selected  \\
			\cline{3-8}
			No. &  Type & Train & Test & Train & Test & Train & Test & Complexity & Features \\
			\midrule
			7 & FNT      & 2.5075 & 2.6481 & 0.88 & 0.88 & 0.0415 & 0.0403 & 16 & 1, 2, 3, 4\\
			8 &          & 2.6030 & 2.6792 & 0.88 & 0.87 & 0.0213 & 0.0217 & 17 & 2, 3, 4\\
			\midrule
			9 & REP-Tree & 2.4460 & 3.6691 & 0.89 & 0.77 & - & - & 51 & 1, 2, 3, 4\\
			10 & MLP      & 2.7698 & 3.8730 & 0.86 & 0.76 & - & - & - & 1, 2, 3, 4\\
			11 & GP Reg. & 2.7978 & 3.7869 & 0.87 & 0.75 & - & - & - & 1, 2, 3, 4\\
			\bottomrule
			\multicolumn{10}{l}{\scriptsize \textbf{Note:} Model Nos. are continued from Table~\ref{tab_die_10FCV_res}}\\
		\end{tabular}			
	\end{table}	
	
	\begin{figure}
		\centering
		\subfigure[Model No. 1]{
			\includegraphics[width=0.45\textwidth]{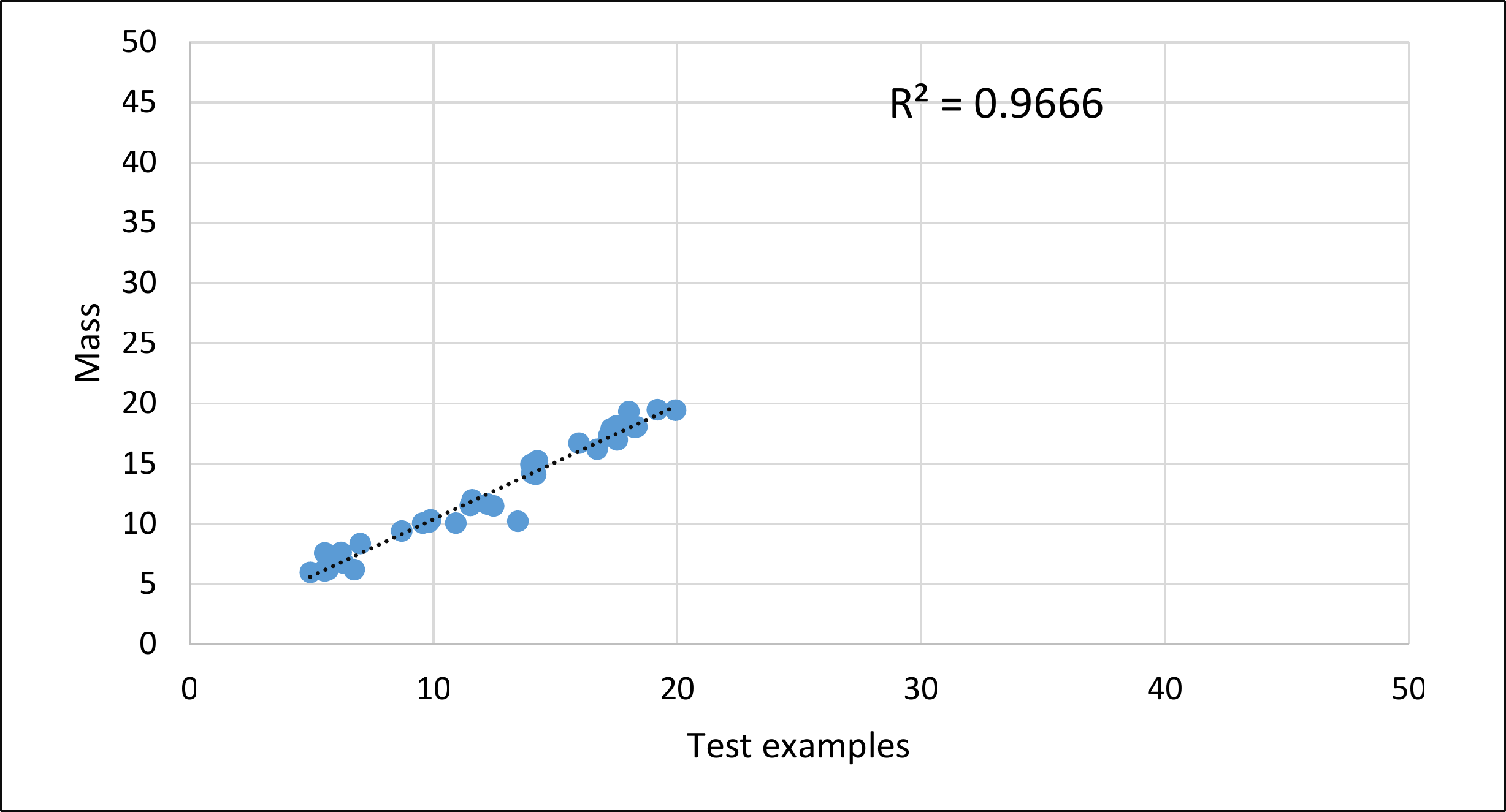}
			\label{fig_die_10FCV_res_a}
		}
		\subfigure[Model No. 1]{
			\includegraphics[width=0.45\textwidth]{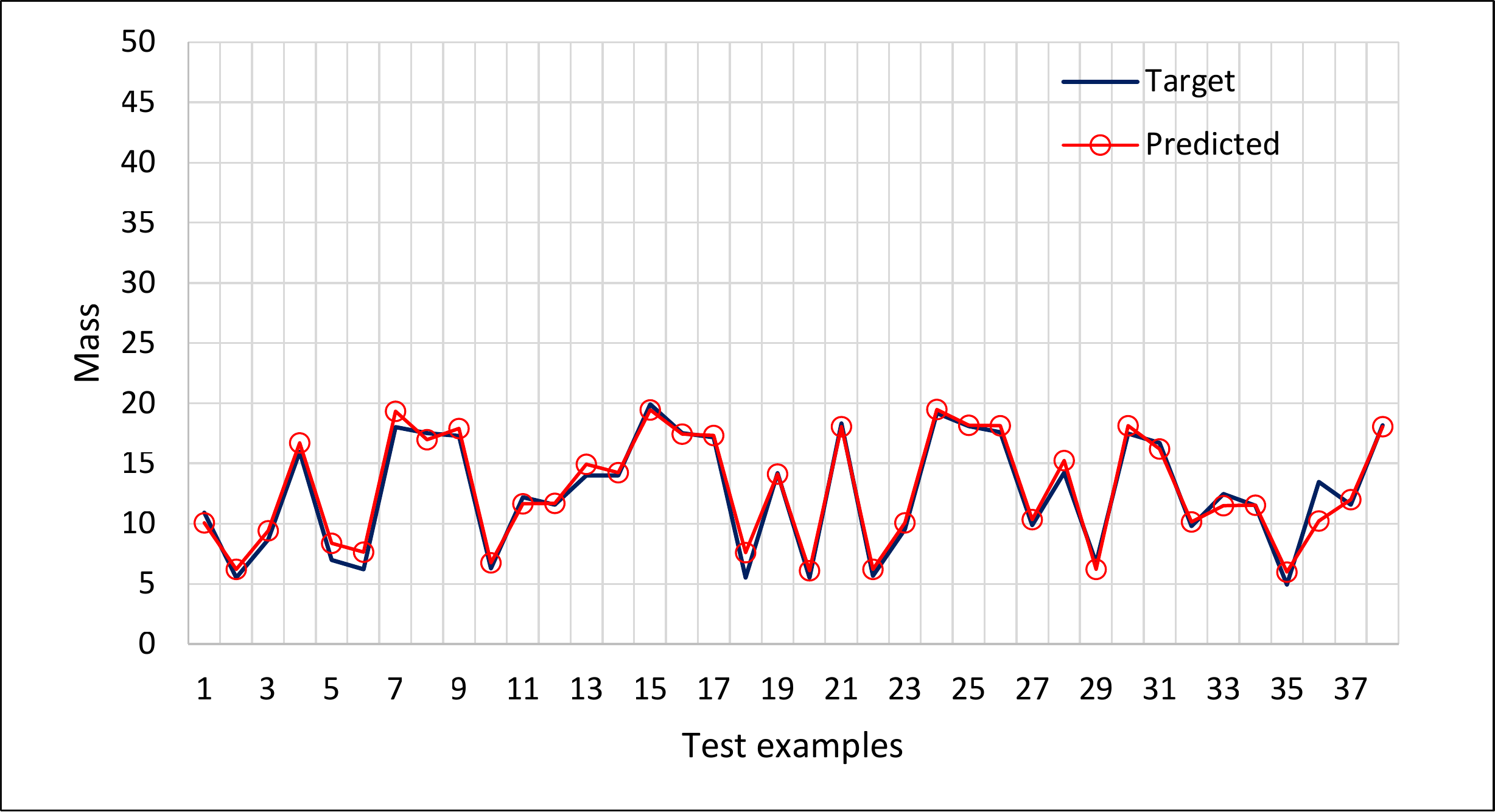}
			\label{fig_die_10FCV_res_b}
		}
		\subfigure[Model No. 2]{
			\includegraphics[width=0.45\textwidth]{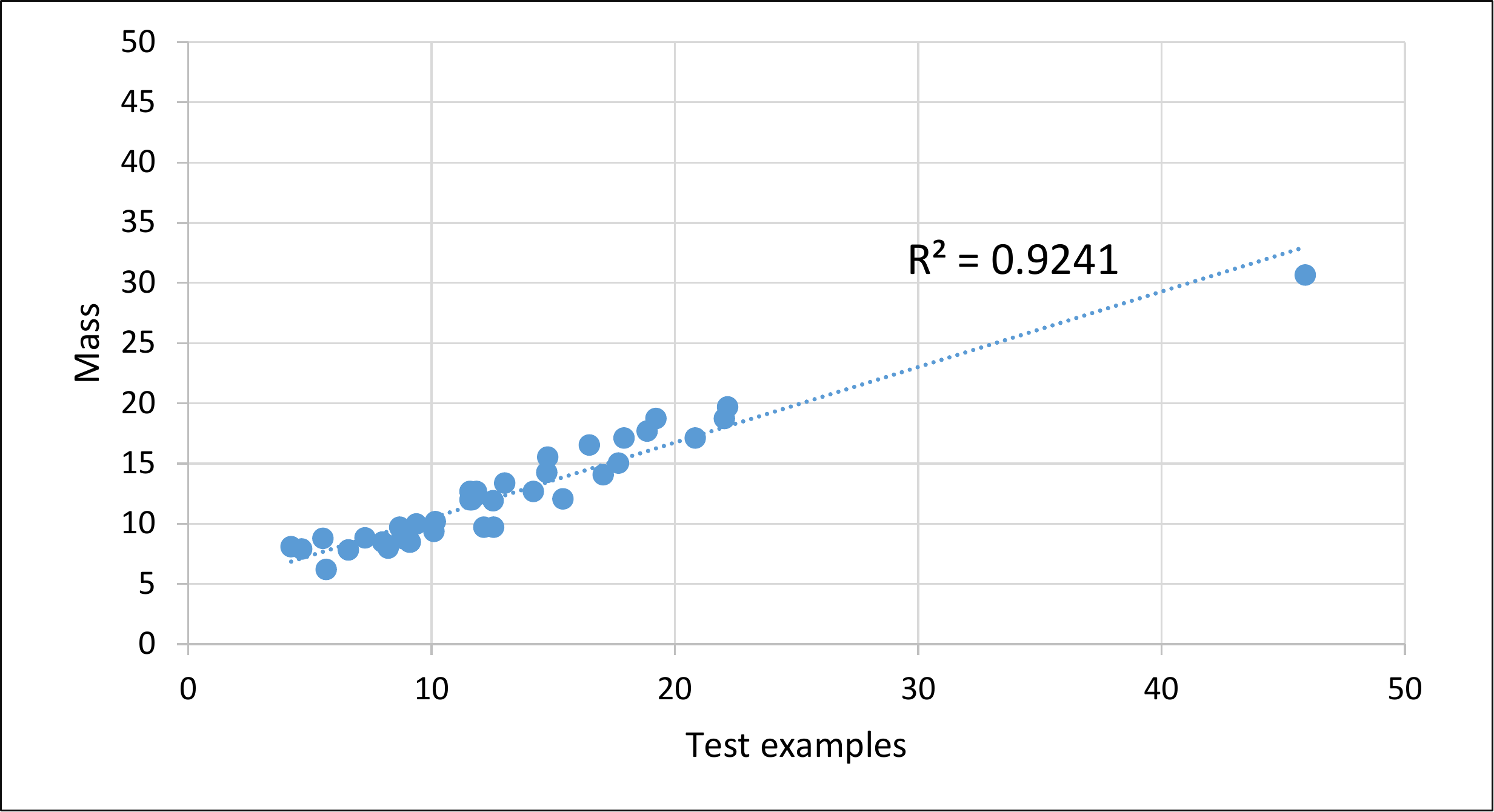}
			\label{fig_die_10FCV_res_c}
		}
		\subfigure[Model No. 2]{
			\includegraphics[width=0.45\textwidth]{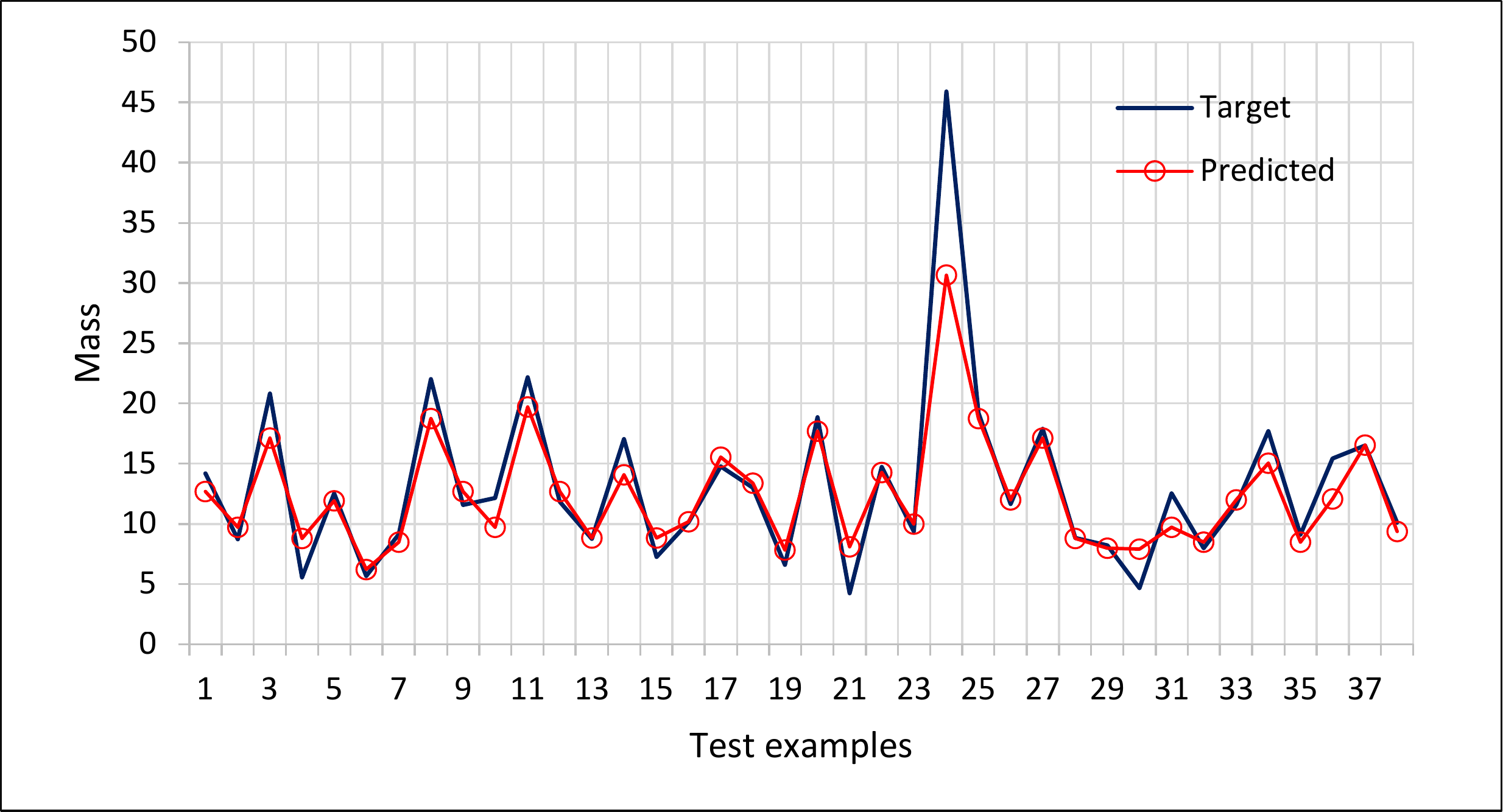}
			\label{fig_die_10FCV_res_d}
		}
		\subfigure[Model No. 3]{
			\includegraphics[width=0.45\textwidth]{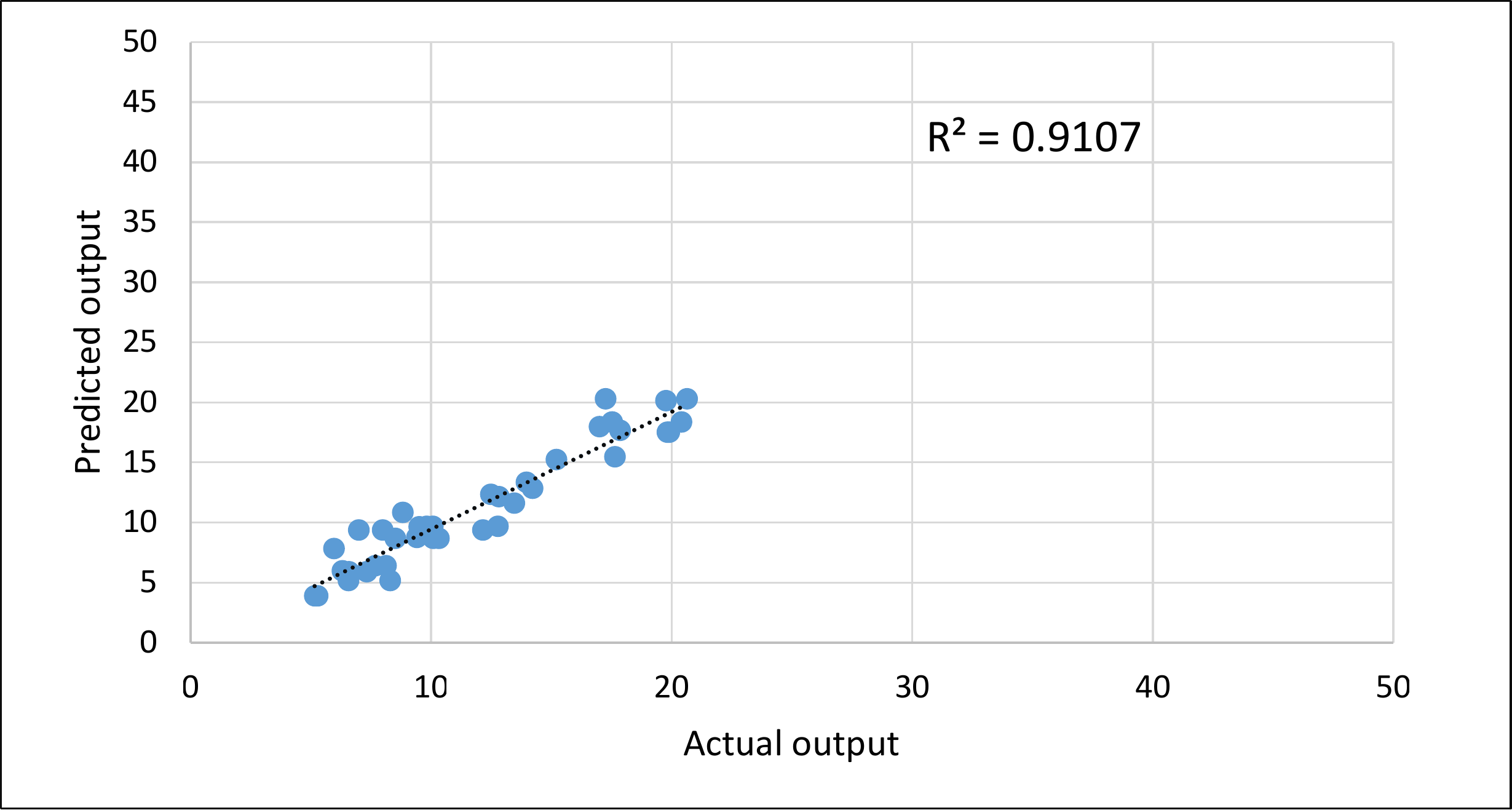}
			\label{fig_die_10FCV_res_e}
		}
		\subfigure[Model No. 3]{
			\includegraphics[width=0.45\textwidth]{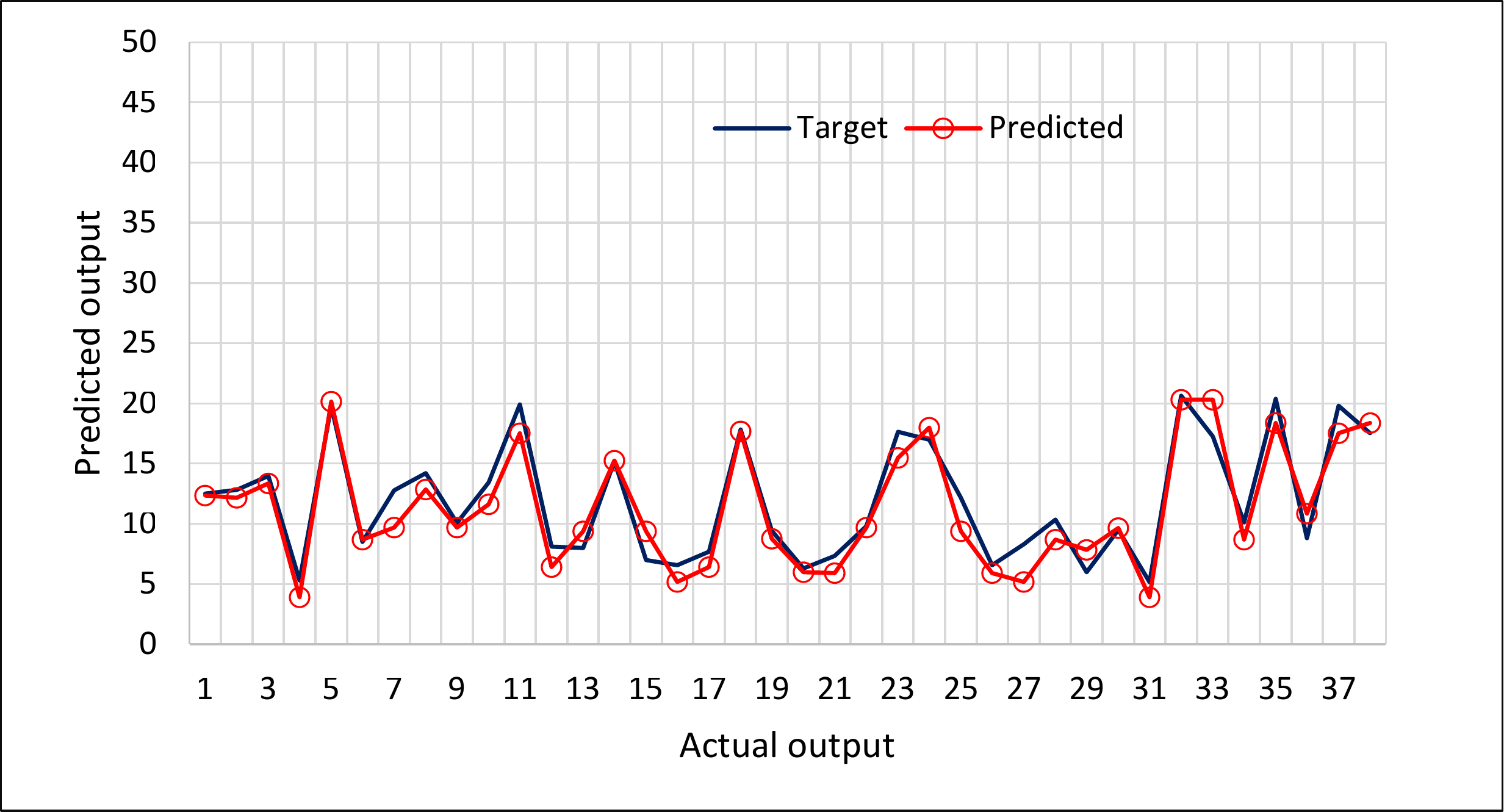}
			\label{fig_die_10FCV_res_f}
		}
		\caption{Models evaluation on unknown test samples: The regression plots (a), (c), and (e) indicates a high correlation between actual and predicted values. The plots (b), (d), and (f) shows the one-to-one mapping of target and prediction of the best models Nos. 1, 2, and 3 (see Table~\ref{tab_die_10FCV_res}). The $ R^2 $ is the squared value of correlation coefficient $ r $, where $ R^2 $ equal to one is the best performance and $ R^2 $ equal to zero is the worst performance.}
		\label{fig_die_10FCV_res}
	\end{figure}
	
	In Fig. 6, each model Nos. 7, and 8, respectively was tested over 50\% of test samples, i.e., 194 randomly chosen test samples, and the scattered plot and the target versus prediction plot were analyzed. Similar to the trend as observed in Fig.~\ref{fig_die_10FCV_res}, the trend observed in Fig.~\ref{fig_die_5x2FCV_res} says that the prediction curve follows the target curve. However, the lower values and the outliers were out of the reach of the prediction curve. The outlier is clearly visible in Fig. 6d. The outlier in our dataset was because of noise or bad value observed during die filling process.
	
	Fig.~\ref{fig_die_5x2FCV_model_a} and Fig.~\ref{fig_die_5x2FCV_model_b} are the illustrations of the created models, where the root node of the tree indicates the output of the models and the leaf nodes (Square boxes with numbers) indicates the input features. In Fig.~\ref{fig_die_5x2FCV_model_a} and Fig.~\ref{fig_die_5x2FCV_model_b}, the input feature $ x_1 $, $ x_2 $, $ x_3 $, and $ x_4 $ indicate the features true density, d50, granule size, and shoe speed, respectively.
	
	\begin{figure}
		\centering
		\subfigure[Model No. 7]{
			\includegraphics[width=0.45\textwidth]{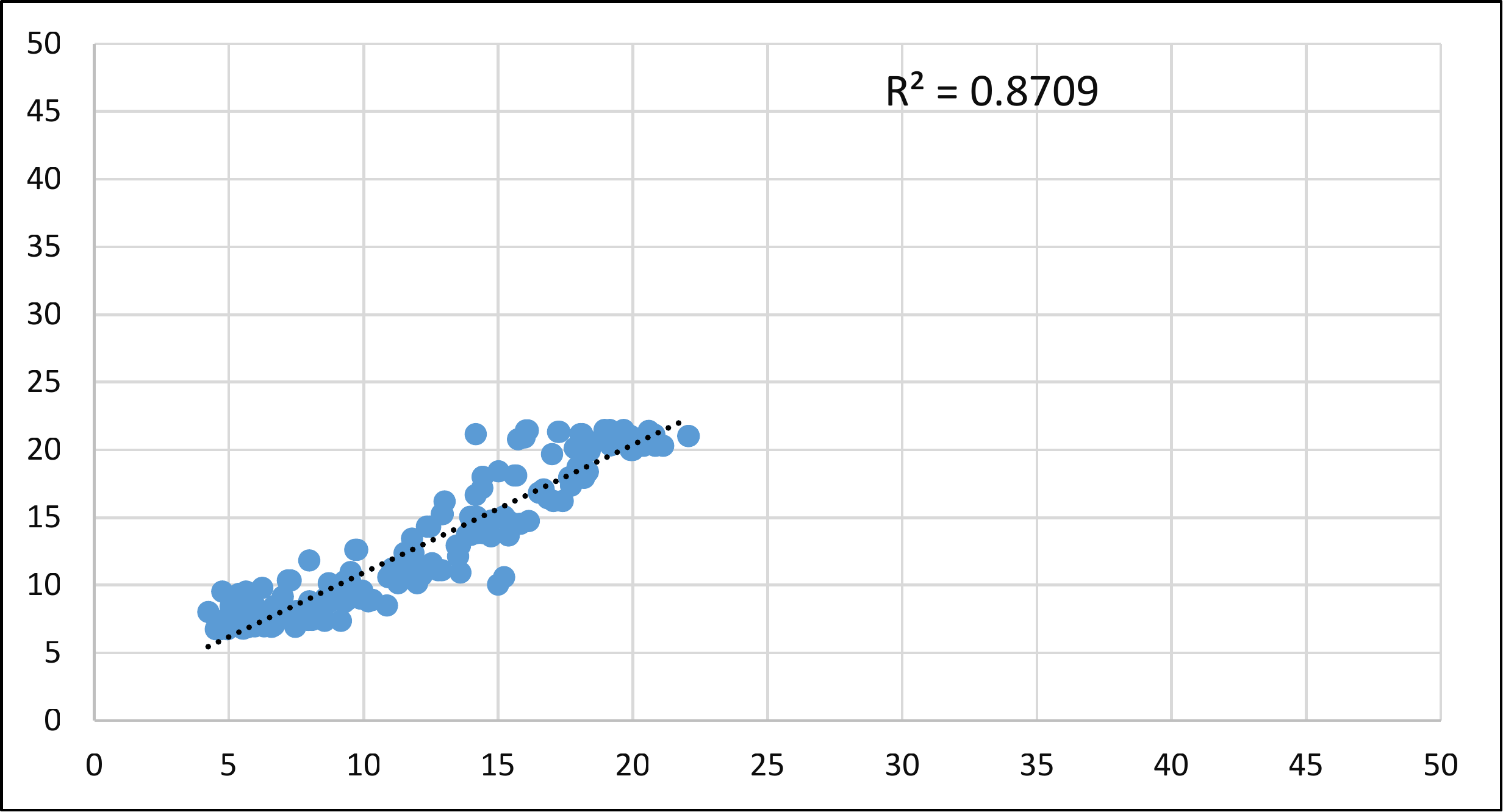}
			\label{fig_die_5x2FCV_res_a}
		}
		\subfigure[Model No. 7]{
			\includegraphics[width=0.45\textwidth]{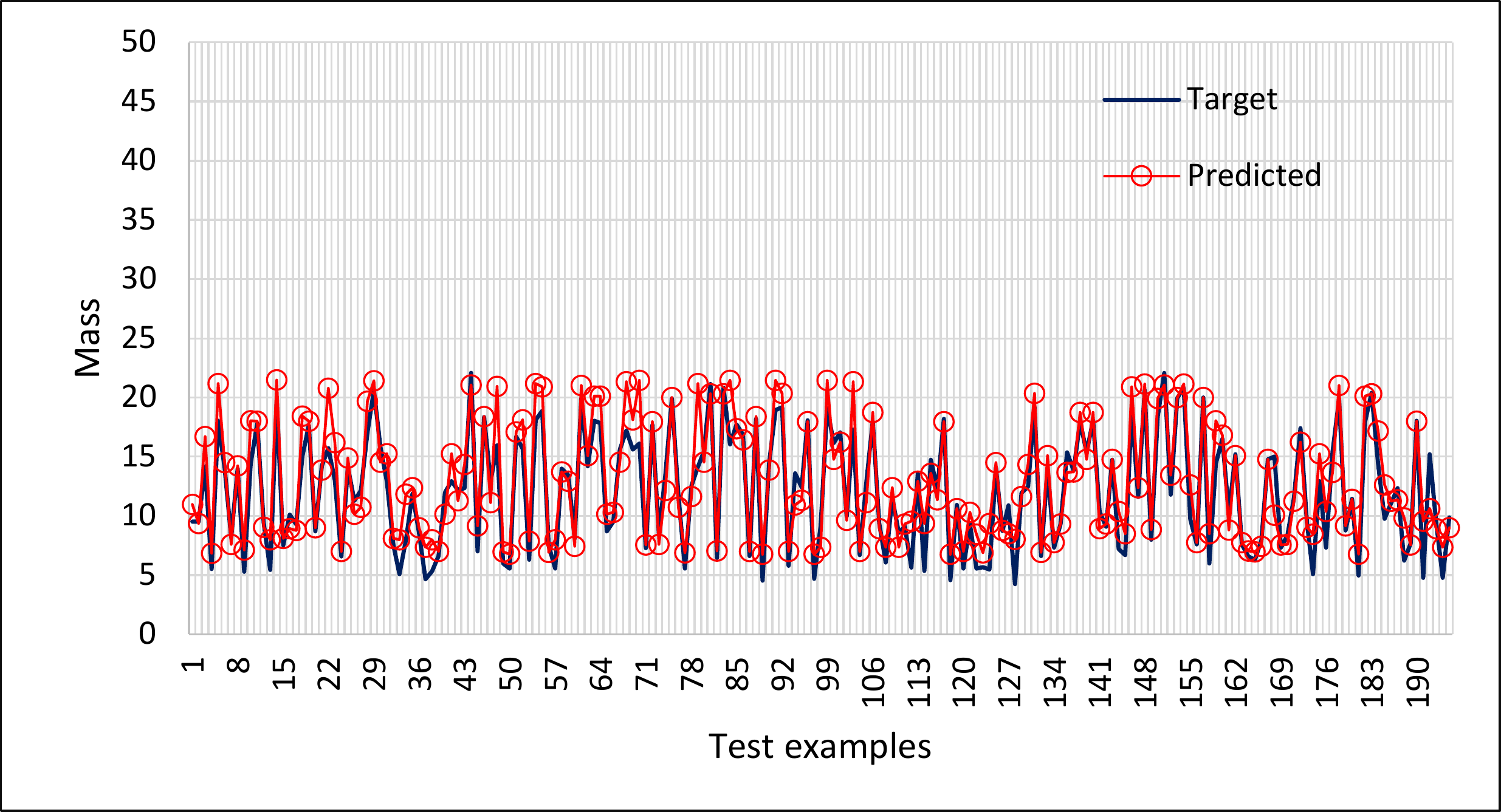}
			\label{fig_die_5x2FCV_res_b}
		}
		\subfigure[Model No. 8]{
			\includegraphics[width=0.45\textwidth]{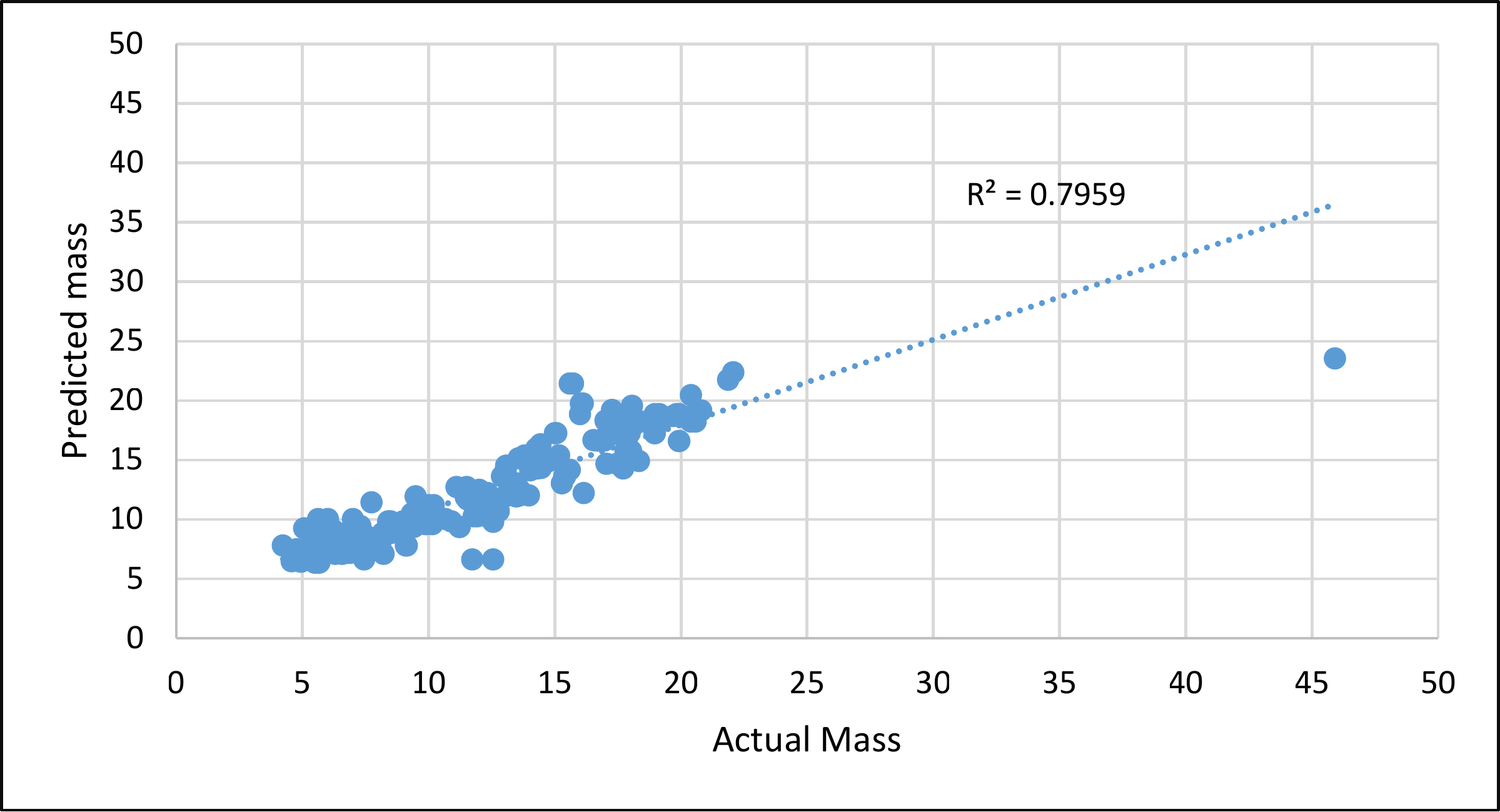}
			\label{fig_die_5x2FCV_res_c}
		}
		\subfigure[Model No. 8]{
			\includegraphics[width=0.45\textwidth]{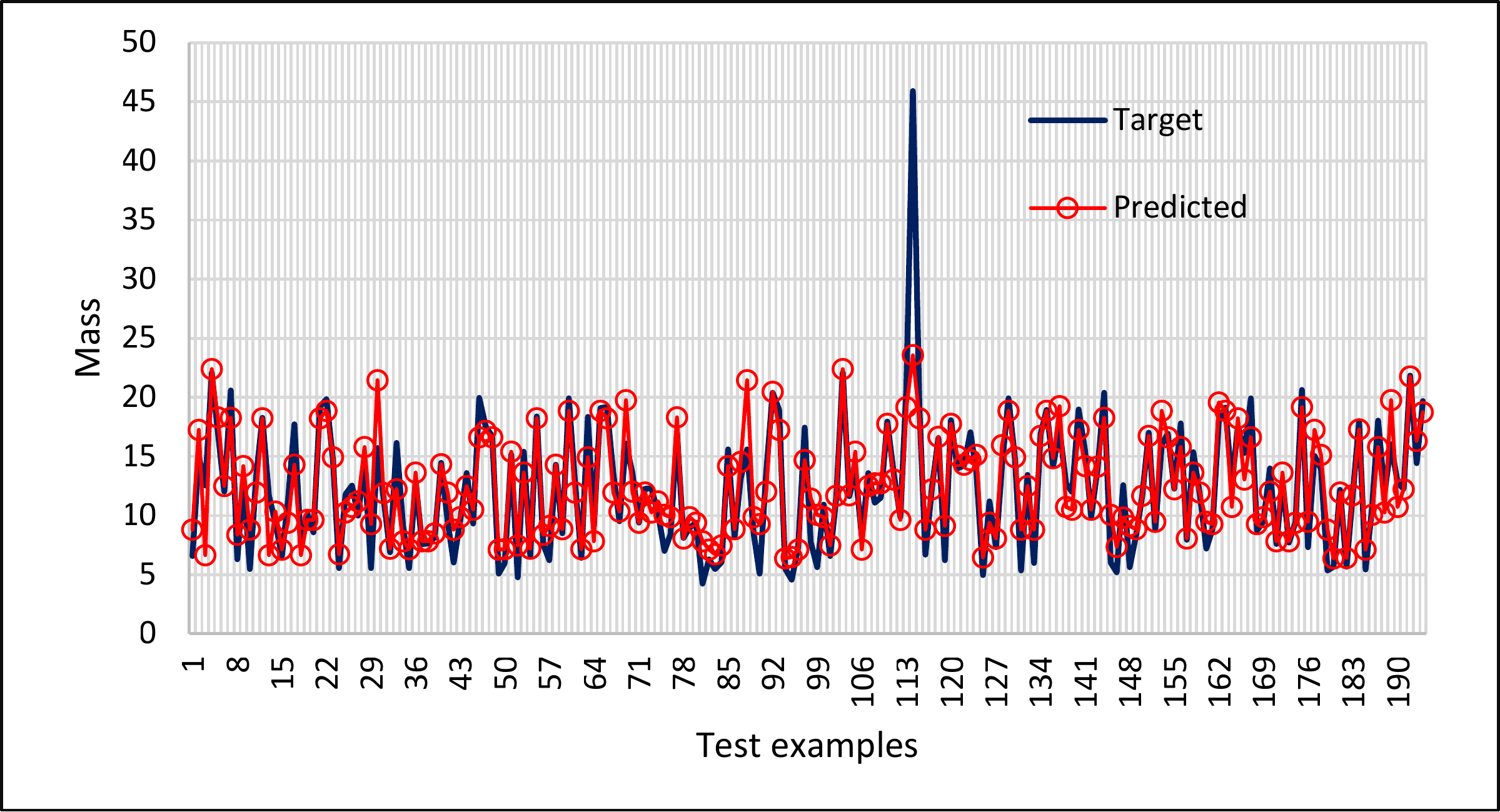}
			\label{fig_die_5x2FCV_res_d}
		}	
		\caption{\scriptsize Models evaluation on unknown test samples. The regression plots (a) and (c) indicates a high correlation between actual and predicted values and plots (b) and (d) shows the one-to-one mapping of target and prediction of the best model Nos. 7 and 8 (Table~\ref{tab_die_5x2FCV_res}). The $ R^2 $ is the squared value of correlation coefficient r, where $ R^2 $ equal to one is the best performance and R2 equal to zero is the worst performance.}
		\label{fig_die_5x2FCV_res}
	\end{figure}
	
	\begin{figure}
		\centering
		\includegraphics[width=0.6\textwidth]{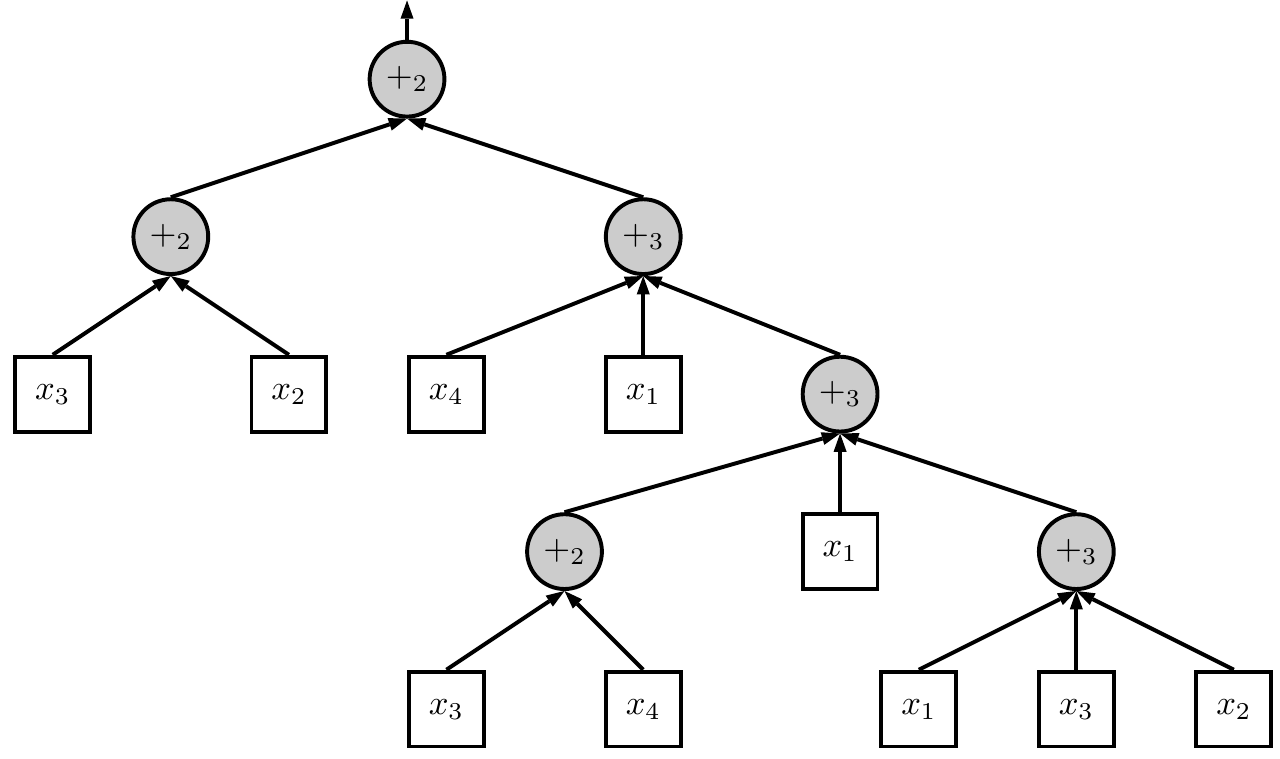}
		\caption{Tree-like structure of predictive model No. 7 created using 5x2FCV method: Complexity equal to 16 is the sum of the computational nodes (node in circles) and the leaf nodes (node in square).}
		\label{fig_die_5x2FCV_model_a}
	\end{figure}
	
	\begin{figure}
		\centering
		\includegraphics[width=0.6\textwidth]{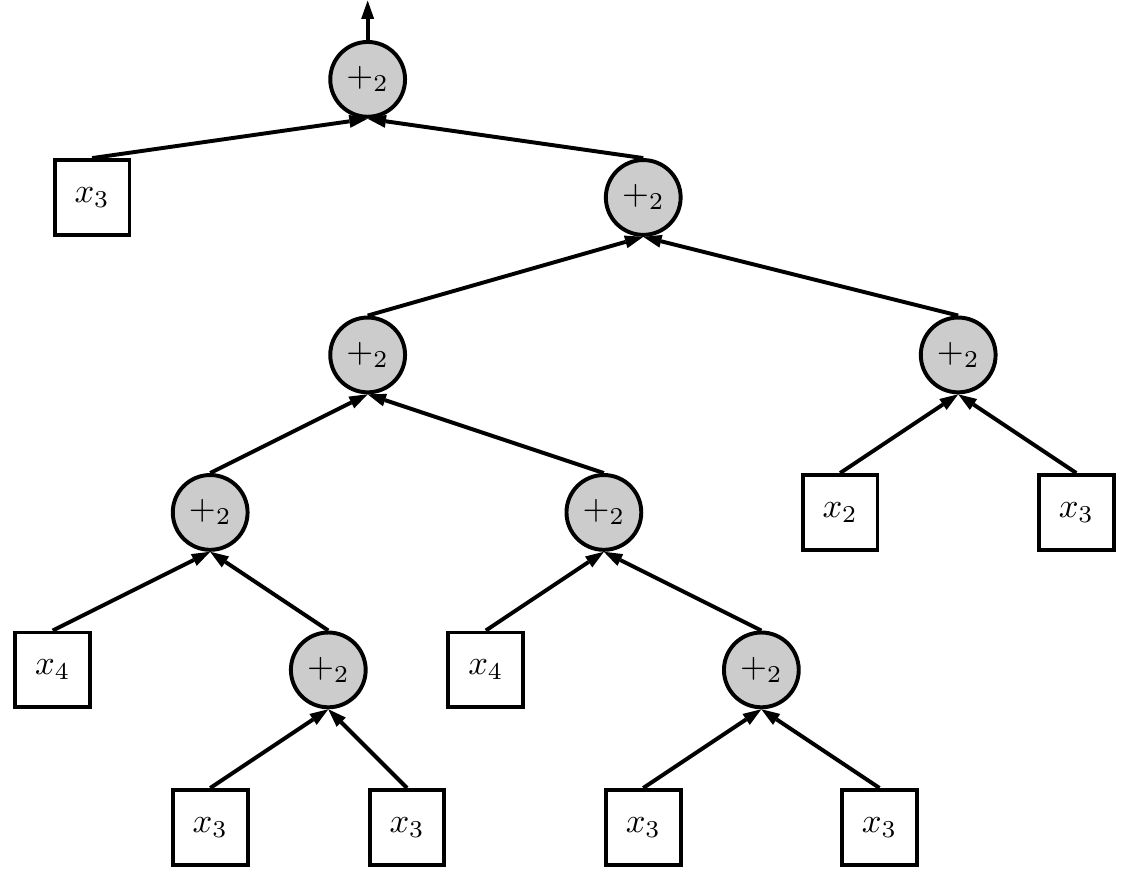}
		\caption{Tree-like structure of predictive model No. 8 created using 5x2FCV method: Complexity equal to 17 is the sum of the computational nodes(node in circles) and the leaf nodes (node in square).}
		\label{fig_die_5x2FCV_model_b}
	\end{figure}	
	\subsubsection{Comparison between 10FCV and 5x2FCV methods}
	Each 10FCV and 5x2FCV methods have their advantages. This is the reason both methods were used for creation and validation of in this work. In 10FCV, a model uses a large sample for training. Thus, possess higher representativeness of real world (data samples) during learning, i.e., the model is trained efficiently. Whereas, in 5x2FCV a model used an equal proportion of data samples for training and testing (smaller training samples than 10FCV, but larger test samples than 10FCV). Thus, 5x2FCV possess higher generalization ability. Hence, at one hand, if a high test correlation coefficient obtained using 10FCV, it indicates that an effective predictive model can be created from the given dataset. Whereas, if a high test correlation coefficient obtained using 5x2FCV, it indicates that a general predictive model can be created from the given dataset.  
	
	The models 7 and 8 were created using 5x2FCV, and their test correlation values were found competitive to the model No. 1using 10FCV method. From our 5x2FCV results, it was found that the model Nos. 7 and 8 were simple in comparison to model Nos. 1, 2, and 3, but their accuracies (correlation coefficient) were slightly poorer in comparison. However, the model Nos. 7 and 8 were tested over 50\% test samples. Hence, it describes that the deposited mass can be efficiently predicted by using die filling process variables knowledge. Moreover, choice of model is subjective. Since simple models possess higher generalization ability, the competitive accuracies of the model Nos. 7 and 8 to the model Nos. 1, 2, and 3 tells that one had to choose between the accuracies and the generalization ability of the models. 
	Fig.~\ref{fig_die_10FCV_res} and Fig.~\ref{fig_die_5x2FCV_res} present the graphical visualization of the 10FCV and 5x2FCV models, respectively, where, the test performance of the models over each test samples were examined. The 5x2FCV models produced similar trend when the models were tested over 50\% of samples. However, the 5x2FCV models (see Fig.~\ref{fig_die_5x2FCV_res_d}) did not predict the outlier as close as 10FCV models (see Fig.~\ref{fig_die_10FCV_res_d}) predicted it.
	
	\subsection{Feature analysis}
	A total 30 models were created using FNT for feature analysis. Since the evolutionary process was used during model creation, the created models selected the input features set that had the highest predictability. Therefore, the RMSEs and selected input feature set by the models were placed into a list. Subsequently, a comprehensive feature analysis was performed. For this purpose, two performance measure dimensions were adopted: feature selection rate $ R $ as defined in~\eqref{eq_sel_rate} and feature predictability score $ P $ as defined in~\eqref{eq_pred_soce_final}. The feature analysis was categorized into two phases. 
	\begin{enumerate}[1)]
		\item 	The identification of individual input features. Here, for~\eqref{eq_sel_rate} and~\eqref{eq_pred_score} the feature set $ |A_j| $ was set to one, which indicates that only one input feature was analyzed at a time. Since there were four input features in our dataset, in this phase, $ P \subset A  $, i.e., $ P $ was equal to  $ \{A_1,A_2,A_3,A_4\} $ (see Table~\ref{tab_die_ind_feture_res} for the definition of $ A_1,\ldots,A_4 $), i.e., $ z $ in~\eqref{eq_pred_soce_final} was equal to four.
		
		\item The identification of feature subset, i.e., identification of the best combination of input feature. Here, for~\eqref{eq_sel_rate} and~\eqref{eq_pred_score} the feature set $ |A_j| $ can be one or two or three or four. After examining the selected feature by the models in the list, there were six different input feature subsets was found. Hence, in this phase, $ Q \subset A  $, i.e., $ Q $ was equal to  $ \{A_1,A_2,A_3,A_4,A_5,A_6\} $ (see Table~\ref{tab_die_set_feture_res} for the definition of $ A_1,\ldots, A_6 $), i.e., $ z $ in~\eqref{eq_pred_soce_final} was equal to six.
	\end{enumerate}
	\subsubsection{Identification of the significance of individual input features}
	Table~\ref{tab_die_ind_feture_res} describes the feature analysis results performed for the individual input features. The significance of individual features true density, d50, granule size, and shoe speed was examined. The features true density and d50 represents the powder properties. Whereas, the granule size and shoe speed represents the die filling process variables. It can be observed that the selection rate and predictability score of d50 (0.62069 and 0.58626) were higher than that of the selection rate and predictability score of true density (0.55173 and 0.54136). Therefore, d50 possess comparatively higher importance as a powder property than that of the true density. Similarly, the process variable granule size was more influential than that of shoe speed. However, the difference of significance level was marginal, but when the entire four input variables were compared, the process variables were having significantly higher selection rate and predictability score than that of the properties of the powder. This fact was also evident from feature subset analysis.
	
	\begin{table}
		\centering
		\caption{Significance of individual input features.}
		\label{tab_die_ind_feture_res}
		\footnotesize 
		\begin{tabular}{llrr}
			\toprule
			\# & Input Features set  & Selection Rate ($ R $) & Predictability Score ($ P $)\\
			\midrule
			1 & $ A_1 $= {True density} & 0.55173 & 0.541356\\
			2 & $ A_2 $= {d50} & 0.62069 & 0.586262\\
			3 & $ A_3 $= {Granule size} & 1 & 1\\
			4 & $ A_4 $= {Shoe speed} & 0.86207 & 0.92563\\
			\bottomrule
		\end{tabular}
	\end{table}	
	\subsubsection{Identification of the best set input features.}
	Table~\ref{tab_die_set_feture_res} shows interesting findings, where it can also observe that the predictability score of subset $ A_4 $ (process variable granule size and shoe speed) and the subsets $ A_1 $, $ A_2 $ and $ A_3 $ (with both process variables combined with one or two powder properties) were higher compared to the subsets $ A_5 $ and $ A_6 $ (subsets where one of the process variables was not used for prediction).
	 
	The subset analysis produced a clear picture. It says that although the predictability score of subset $ A_3 $ was highest, the selection rate of subset $ A_1 $ was highest. This indicates that the evolutionary process often preferred to use the combination of entire features, i.e., the set $ A_1 $. However, among the subsets $ A_1 $, $ A_2 $, and $ A_3 $, the selection rate of subset $ A_2 $ was higher, which indicates d50 had the higher ability to represent powder properties than that of true density, but again the difference was marginal (say approximately higher by only four percent).
	
	\begin{table}
		\centering
		\caption{Optimal subset of input features.}
		\label{tab_die_set_feture_res}
		\footnotesize 
		\begin{tabular}{llrr}
		\toprule
		\# & Input Feature set & Selection Rate ($ R $) & Predictability Score ($ P $)\\
		\midrule
		1 & $ A_1 $= {True density, d50, Granule size, Shoe speed}   & 0.31035 & 0.969497\\
		2 & $ A_2 $= {d50, Granule size, Shoe Speed} & 0.17242 & 0.941601\\
		3 & $ A_3 $= {True density, Granule size, Shoe speed} & 0.13793 & 1\\
		4 & $ A_4 $= {Granule size, Shoe speed} & 0.24138 & 0.979663\\
		5 & $ A_5 $= {True density, d50, Granule size} & 0.10345 & 0.493741\\
		6 & $ A_6 $= {d50, Granule size} & 0.03448 & 0.470451\\
		\bottomrule
	\end{tabular}
	\end{table}	
	
	\subsubsection{Accuracy of Model Predictions}
\begin{figure}
		\centering
		\subfigure[]{
			\includegraphics[width=0.75\textwidth,]{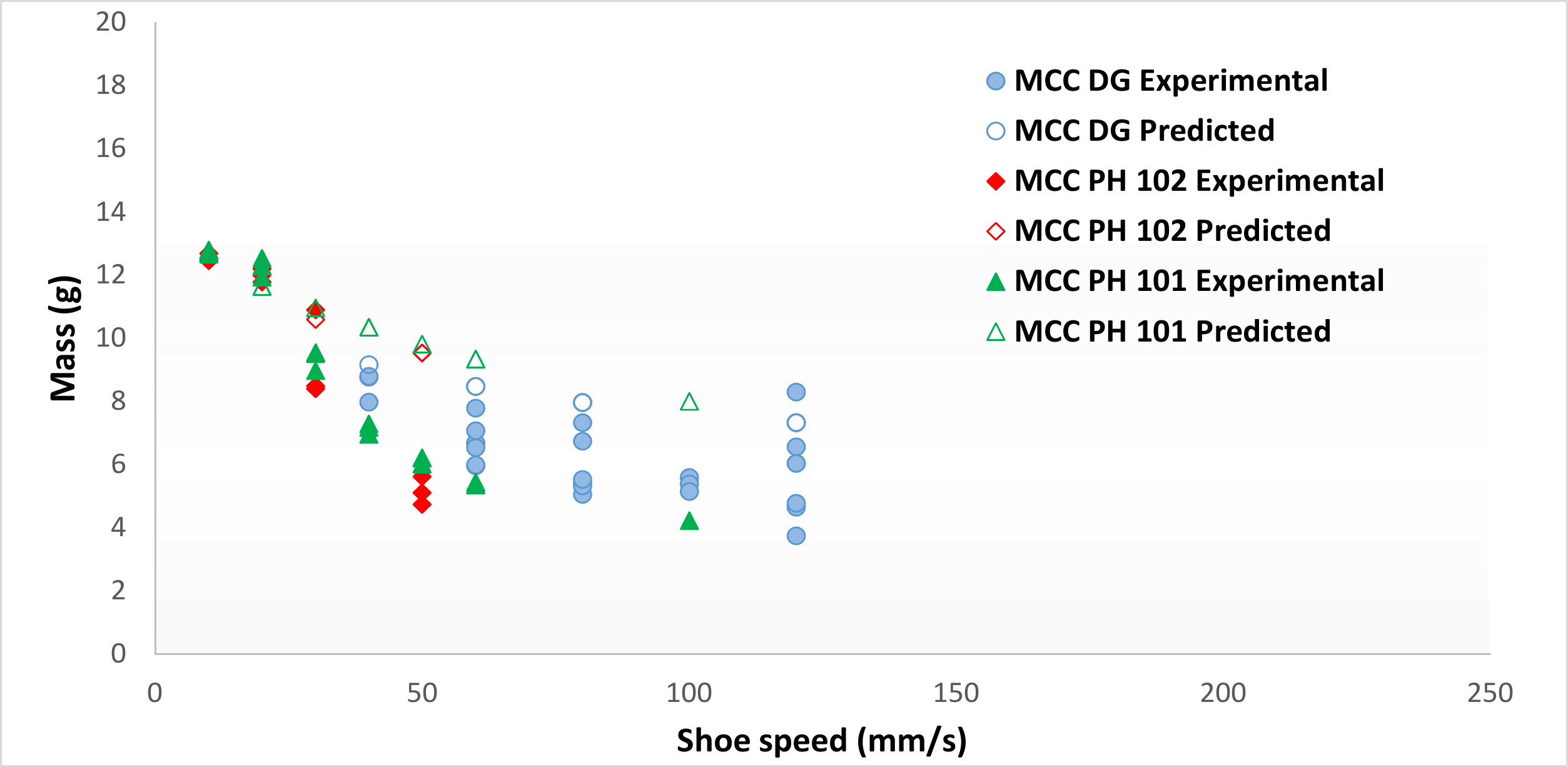}
			\label{fig_die_model_data_com_a}
		}
		\subfigure[]{
			\includegraphics[width=0.75\textwidth]{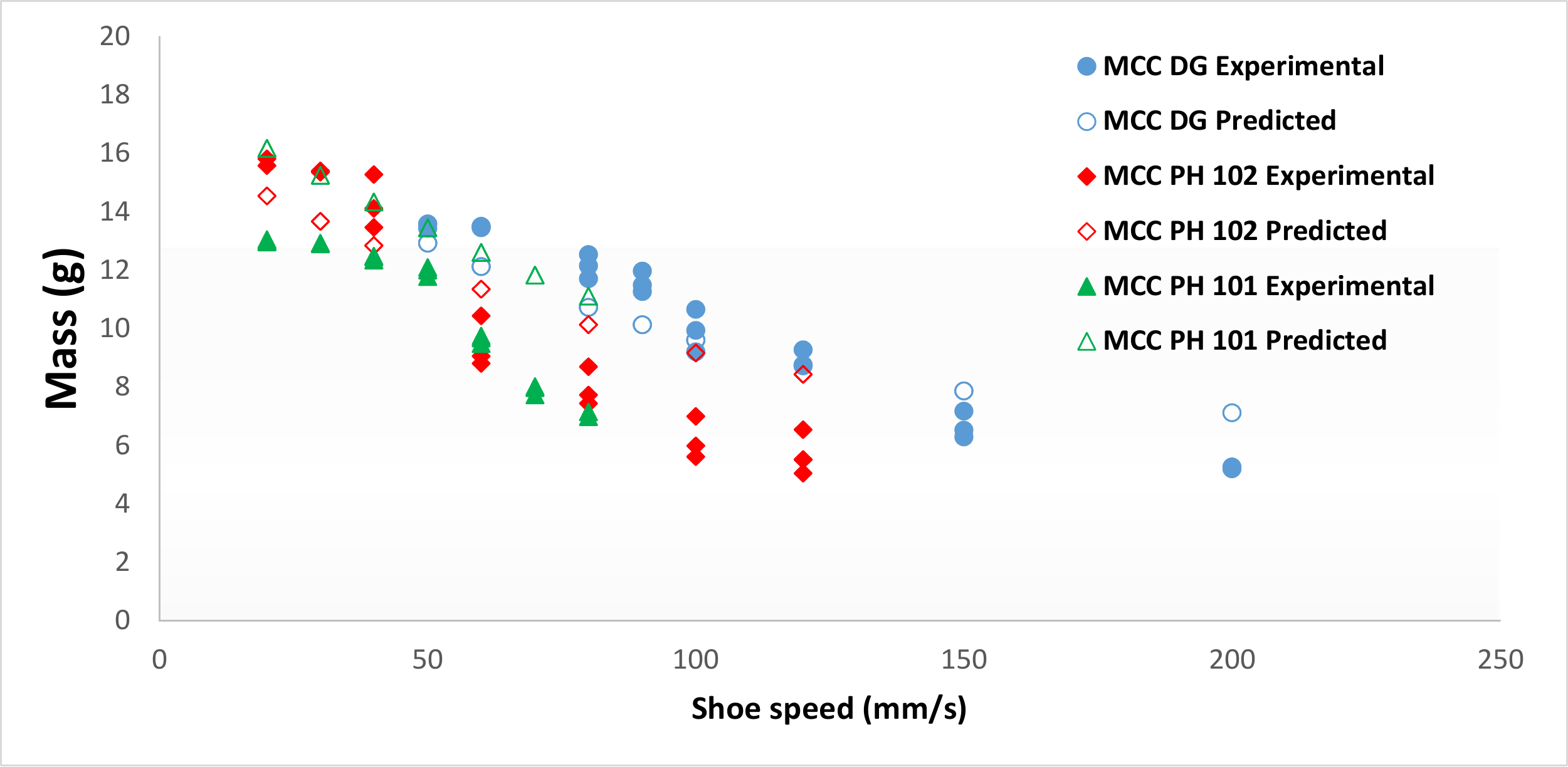}
			\label{fig_die_model_data_com_b}
		}
		\subfigure[]{
			\includegraphics[width=0.75\textwidth]{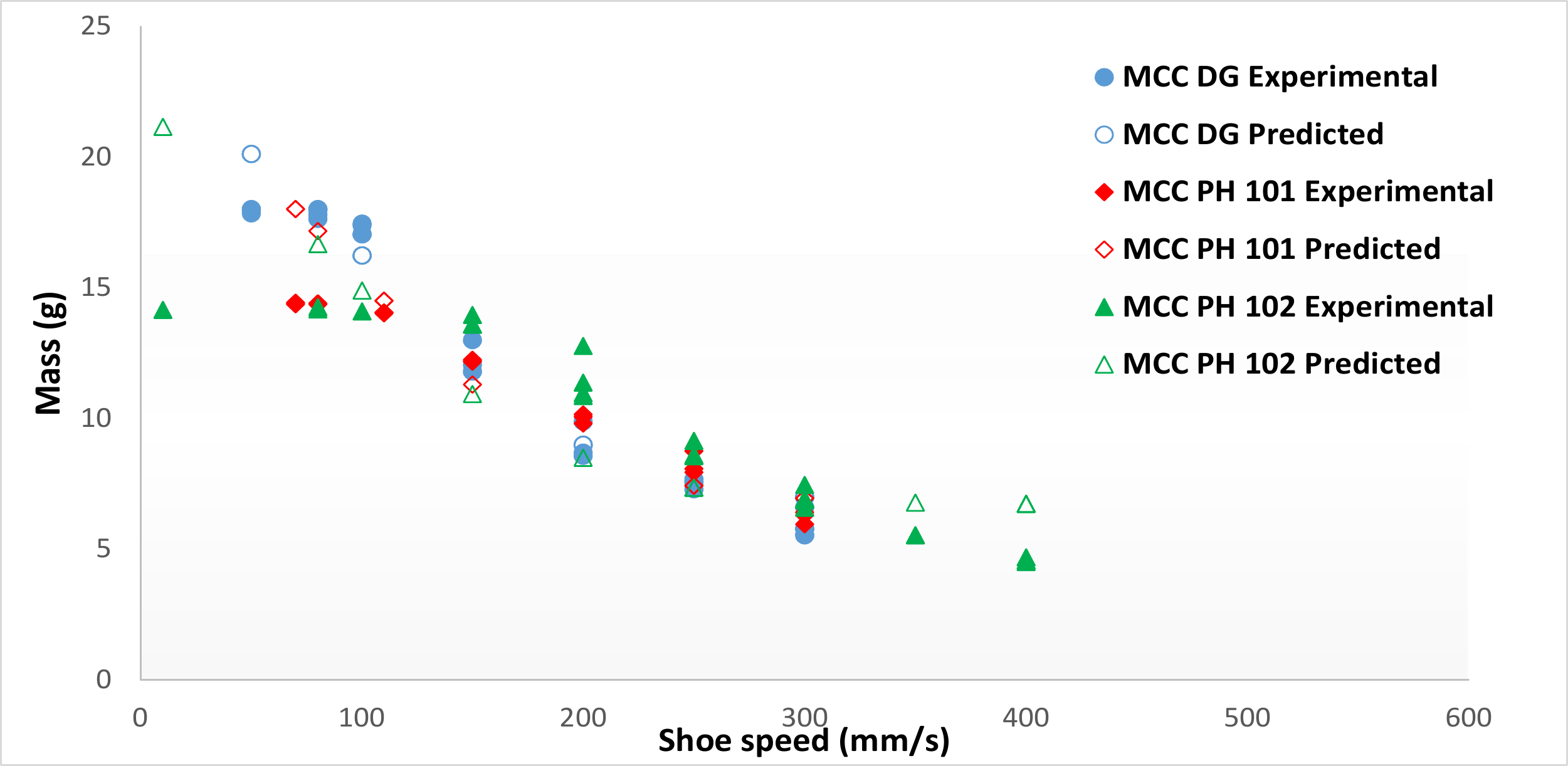}
			\label{fig_die_model_data_com_c}
		} 
		\caption{Comparison of experimental results and model predictions for 3 MCC granules of different size ranges: a) 1-90~\si{\micro\meter}, b) 90-250~\si{\micro\meter}, c) 250-500~\si{\micro\meter}.}
		\label{fig_die_model_data_com_P1}    
	\end{figure}

\begin{figure}	
		\centering
	    \subfigure[]{
	    	\includegraphics[width=0.75\textwidth]{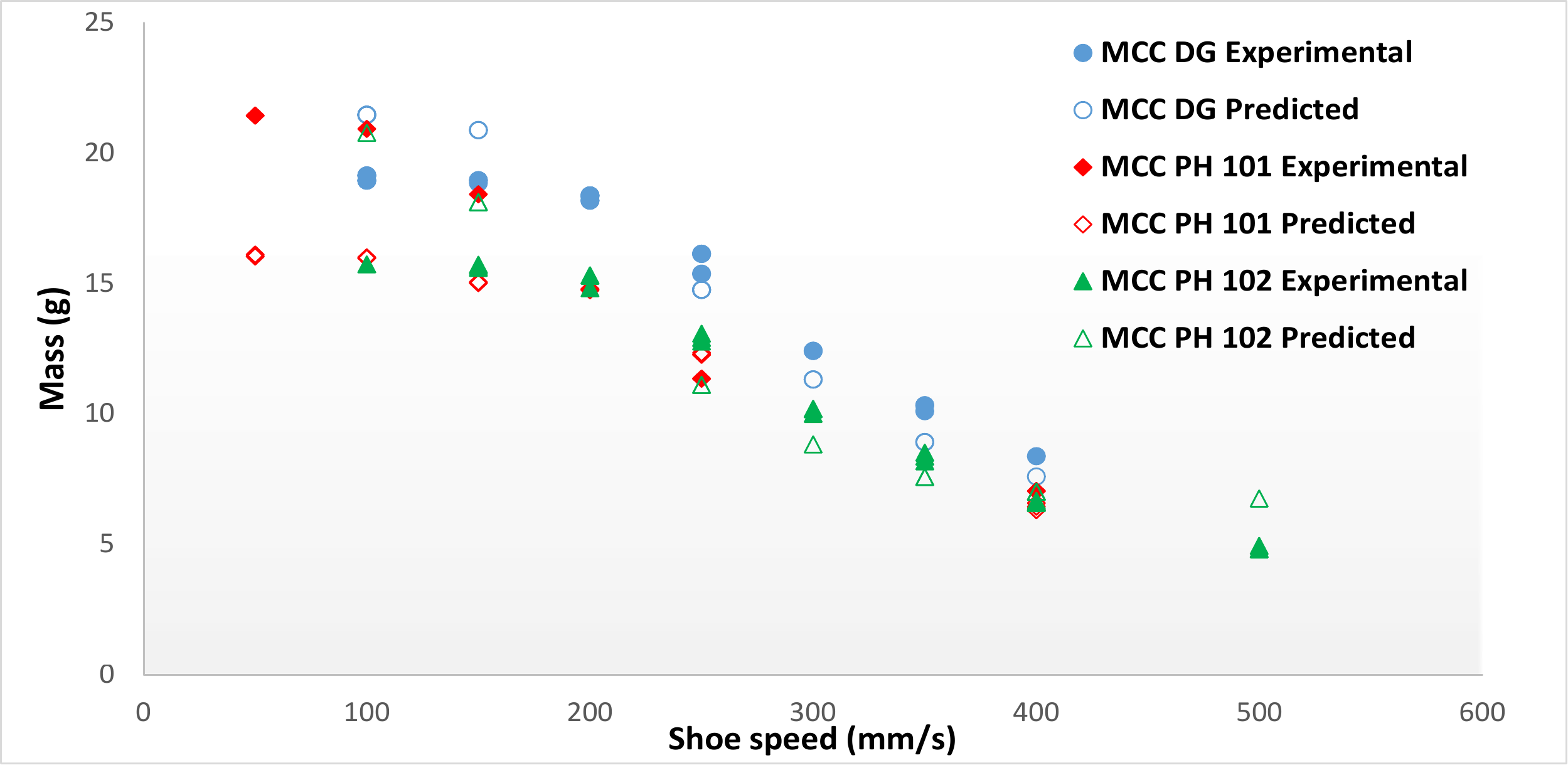}
	    	\label{fig_die_model_data_com_d}
	    }
	    \subfigure[]{
	    	\includegraphics[width=0.75\textwidth]{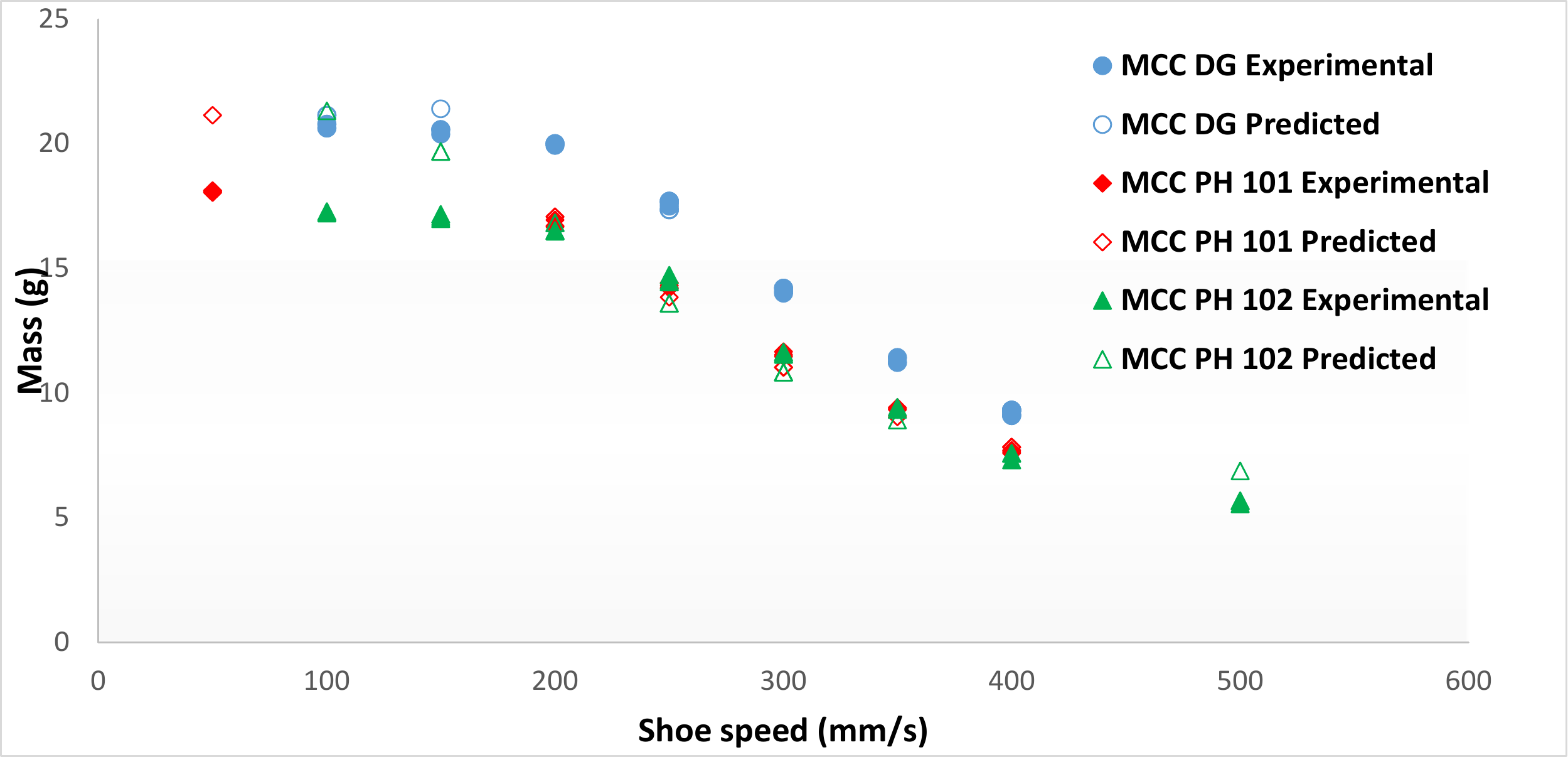}
	    	\label{fig_die_model_data_com_e}
	    }
	    \subfigure[]{
	    	\includegraphics[width=0.75\textwidth]{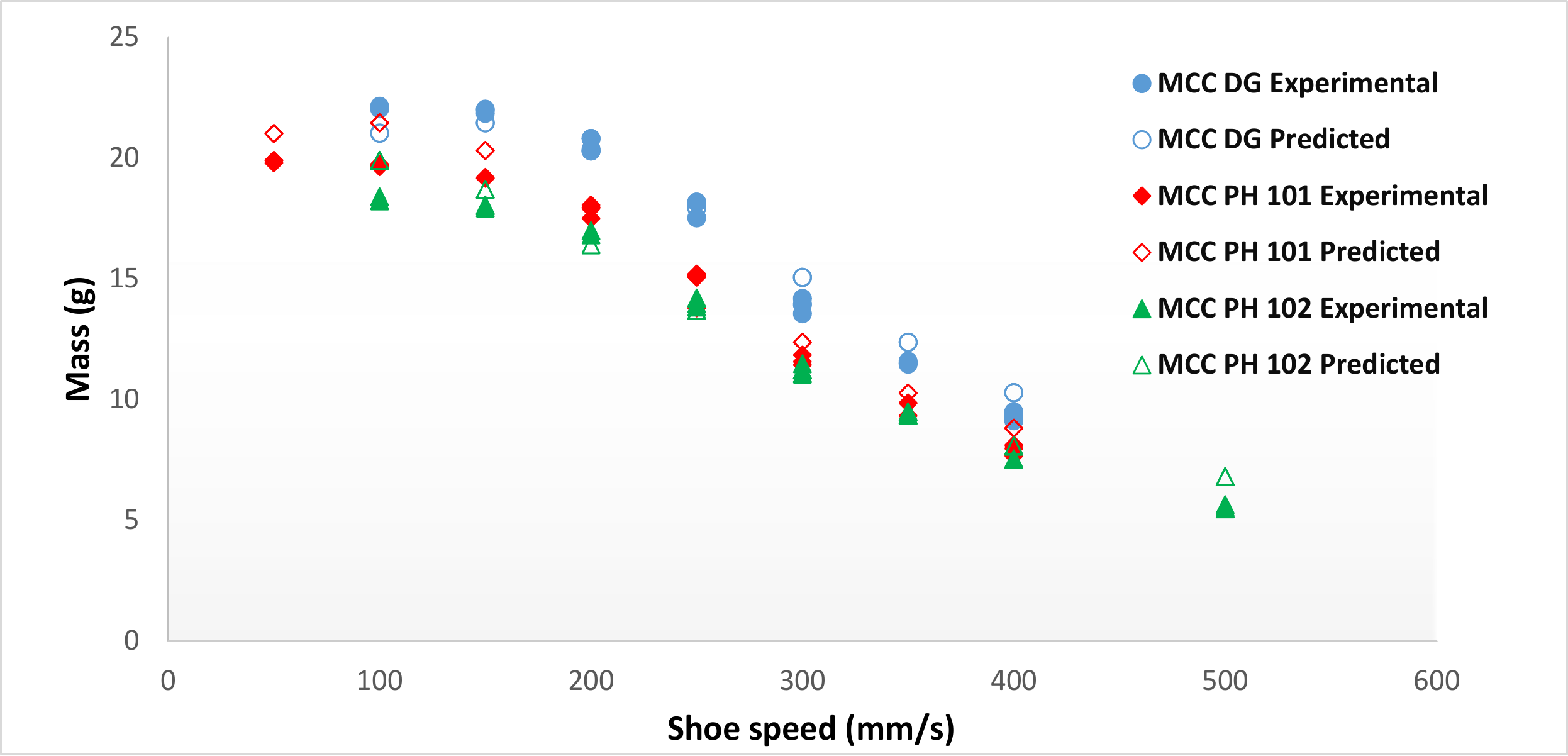}
	    	\label{fig_die_model_data_com_f}
	    }			
		\caption{Comparison of experimental results and model predictions for 3 MCC granules of different size ranges: a) 500-1000~\si{\micro\meter}, b) 1000-1400~\si{\micro\meter}, and c) 1400-2360~\si{\micro\meter}.}
		\label{fig_die_model_data_com_P2}
	\end{figure}

	Fig.~\ref{fig_die_model_data_com_P1} and Fig.~\ref{fig_die_model_data_com_P2} show the comparison between die filling experimental results and predicted results from the model No 1 of the 10FCV method using FNT. This model was chosen because it has the highest value of $ R^2 $, which leads to a better fitting between experimental and predicted data. More specifically, Fig.~\ref{fig_die_model_data_com_P1} and Fig.~\ref{fig_die_model_data_com_P2} present the mass collected after each experiment as a function of the shoe speed of the three different MCCs powders for six different granule size ranges. It shows that there is a decrease of mass deposited into the die with increasing shoe speed for all the materials and granules size ranges investigated. Moreover, a general increase of deposited mass at a consistent shoe speed was found with the increasing granule size. MCC DG tends to have higher mass deposited values compare to MCC PH 102 and MCC PH 101 at all the shoe speeds considered and for all the different size ranges analyzed, exception for the granule size range 250-500~\si{\micro\meter} (Fig.~\ref{fig_die_model_data_com_c}). MCC PH 101 and MCC PH 102 show an identical trend in all the experiments performed. Interestingly, results for finest granules (Fig.~\ref{fig_die_model_data_com_a} and Fig.~\ref{fig_die_model_data_com_b}) appears to have larger variations (due to the lower experimental reproducibility) compare to those for coarser granules (Fig.~\ref{fig_die_model_data_com_e} and Fig.~\ref{fig_die_model_data_com_f}). It is clear that the models give better predictions for coarser granules (Fig.~\ref{fig_die_model_data_com_e} and Fig.~\ref{fig_die_model_data_com_f}) than for finer granules (Fig.~\ref{fig_die_model_data_com_a} and Fig.~\ref{fig_die_model_data_com_b}) for all the materials under investigation. In particular, the model gives almost identical values to the measured ones for coarser granules, which proves that the FNT method used can predict die filling behavior for such materials with high accuracy. The accuracy of the model appears to rely on the consistency (or scattering) of the experimental data.
		
	\section{Conclusions}
	Flexible neural tree (FNT) was used to predict die filling behavior of MCCs granules of different size ranges. Two main methods were investigated, 10-fold cross-validation (CV) and 5x2-fold CV. Computational intelligence (CI) models were developed using FNT, multilayer perceptron, reduced error pruning tree, and Gaussian process regression. It was observed that the flexible neural tree models performed better than other CI techniques. Additionally, by examining FNT models of each method, it was found that 10-fold CV was a more efficient method with a higher correlation coefficient than the 5x2 fold CV. The experimental results were used as inputs and outputs of the FNT models. The constructed model efficiently predicted the deposited mass based on the knowledge gathered from the experimental data. Similarly, the feature analysis discovered that the shoe speed and the granule size are more significant in terms of governing the deposited mass than the raw powder properties (true density and d50). Interestingly, die filling behavior of coarser granules are easier to predict than fine granules for all the materials considered. This is due to the higher reproducibility of the experimental data for larger granules.
	
	\section*{Acknowledgments}
	This work was supported by the IPROCOM Marie Curie Initial Training Network, funded through the People Programme (Marie Curie Actions) of the European Union’s Seventh Framework Programme.
	
	\bibliographystyle{IEEEtran}
	\bibliography{fnt_die_rev_1}   
	
	
\end{document}